\documentclass[dvipsnames,format=manuscript,nonacm]{acmart}
\usepackage{mathtools}
\renewcommand{\vec}{\mathbf}
\usepackage{bm}
\usepackage{subfigure}
\usepackage{epsfig}
\usepackage{paralist}
\usepackage{longtable}
\usepackage{algorithmicx}
\usepackage{colortbl}
\usepackage[linesnumbered,ruled,lined,boxed]{algorithm2e}
\AtBeginDocument{%
  \providecommand\BibTeX{{%
    \normalfont B\kern-0.5em{\scshape i\kern-0.25em b}\kern-0.8em\TeX}}}

\setcopyright{none}
\renewcommand\footnotetextcopyrightpermission[1]{}
\usepackage{geometry}
\begin{document}

\title[Clustering in Dynamic Environments: A Framework for Benchmark Dataset Generation]{Clustering in Dynamic Environments: A Framework for Benchmark Dataset Generation with Heterogeneous Changes}

\author{Danial~Yazdani}
\email{danial.yazdani@gmail.com}
\affiliation{%
  \institution{Faculty of Engineering \& Information Technology, University of Technology Sydney}
  \city{Ultimo}
  \country{Australia}
  \postcode{2007}
}

\author{Juergen~Branke}
\email{Juergen.Branke@wbs.ac.uk}
\affiliation{%
  \institution{Warwick Business school, University of Warwick}
  \city{Coventry}
  \country{United Kingdom}
  \postcode{CV4 7AL}
}

\author{Mohammad~Sadegh~Khorshidi}
\email{msadegh.Khorshidi.ak@gmail.com}
\affiliation{%
  \institution{Faculty of Engineering \& Information Technology, University of Technology Sydney}
  \city{Ultimo}
  \country{Australia}
  \postcode{2007}
}

\author{Mohammad~Nabi~Omidvar}
\email{m.n.omidvar@leeds.ac.uk}
\affiliation{%
  \institution{School of Computing and Leeds University Business School, University of Leeds}
  \city{Leeds}
  \country{United Kingdom}
  \postcode{LS2 9JT}
}

\author{Xiaodong~Li}
\email{xiaodong.li@rmit.edu.au}
\affiliation{%
  \institution{RMIT University}
  \city{Melbourne}
  \country{Australia}
  \postcode{3000}
}

\author{Amir~H.~Gandomi}
\email{Gandomi@uts.edu.au}
\authornote{Corresponding author.}
\affiliation{%
  \institution{Faculty of Engineering \& Information Technology, University of Technology Sydney}
  \city{Ultimo}
  \country{Australia}
  \postcode{2007}
}
\affiliation{%
  \institution{University Research and Innovation Center (EKIK), Obuda University}
  \city{Budapest}
  \country{Hungary}
  \postcode{1034}
}

\author{Xin~Yao}
\email{xinyao@LN.edu.hk}
\affiliation{%
  \institution{Department of Computing and Decision Sciences, Lingnan University}
  \city{Hong Kong}
  \country{China}
}
\affiliation{%
  \institution{CERCIA, School of Computer Science, University of Birmingham}
  \city{Birmingham}
  \country{United Kingdom}
  \postcode{B15 2TT}
}
\renewcommand{\shortauthors}{Yazdani et al.}

\begin{abstract}
\textbf{Abstract:} Clustering in dynamic environments is of increasing importance, with broad applications ranging from real-time data analysis and online unsupervised learning to dynamic facility location problems. While meta-heuristics have shown promising effectiveness in static clustering tasks, their application for tracking optimal clustering solutions or robust clustering over time in dynamic environments remains largely underexplored. This is partly due to a lack of dynamic datasets with diverse, controllable, and realistic dynamic characteristics, hindering systematic performance evaluations of clustering algorithms in various dynamic scenarios. This deficiency leads to a gap in our understanding and capability to effectively design algorithms for clustering in dynamic environments. To bridge this gap, this paper introduces the Dynamic Dataset Generator (DDG). DDG features multiple dynamic Gaussian components integrated with a range of heterogeneous, local, and global changes. These changes vary in spatial and temporal severity, patterns, and domain of influence, providing a comprehensive tool for simulating a wide range of dynamic scenarios.
\end{abstract}




\maketitle

\section{Introduction}
\label{sec:Introduction}

Clustering refers to the process of grouping a set of objects into subsets or `clusters' such that objects within the same cluster are more similar to each other than to those in other clusters~\cite{jain1999data}. 
This unsupervised learning technique is crucial for discovering inherent structures within data and revealing patterns and relationships without prior knowledge of the categories. 
Clustering is widely applicable across various domains, from image and pattern recognition to market segmentation and facility location problems~\cite{xu2005survey}. 

Clustering in dynamic environments significantly amplifies the challenge compared to its static counterpart, as it involves variables that change over time, such as objects, distributions, underlying structures, and the number of clusters. 
This inherent dynamism introduces an additional layer of complexity, necessitating advanced strategies capable of adapting and evolving within this ever-changing environment. 
Clustering in such environments is crucial for a wide array of applications, ranging from real-time data analysis in the presence of concept drift~\cite{moulton2019clustering} and online unsupervised learning~\cite{zhan2020online}, to the organization of various entities such as customers and demands in dynamic facility location problems~\cite{karatas2021dynamic}, as well as the clustering of moving objects like vehicles~\cite{li2022evolutionary}.

Clustering within dynamic environments can be framed as Dynamic Optimization Problems (DOPs), where the goal is to optimally cluster a set of objects based on one or more specified objective functions~\cite{handl2007ant}. 
A time-varying objective function $f^{(t)}$ in a DOP can be described as $f^{(t)}(\vec{x})=f\left(\vec{x},\bm\alpha^{(t)}\right)$, where $t\in [0,t_{\mathrm{max}}]$ denotes the time index, $\vec{x}$ symbolizes a solution within the $d$-dimensional search space, and $\bm\alpha$ is a vector of time-dependent control parameters influencing the objective function. 
In response to environmental changes and the inadequacy of prior solutions, a new, more fitting solution must be deployed.
Solving DOPs, therefore, requires an algorithm not just to locate but also to continually track the desirable solutions over time.
A typical approach in solving DOPs involves restarting the optimization process after each environmental change. 
However, this method proves inefficient for DOPs experiencing rapid change frequencies or when a swift solution deployment is crucial post-change, a common requirement referred to as `quick recovery' in numerous real-world DOPs~\cite{nguyen2011thesis}.

Dynamic optimization algorithms (DOAs) have demonstrated their potential in effectively solving various classes of DOPs~\cite{yazdani2021DOPsurveyPartA,yazdani2018thesis}.
DOAs' distinct advantage lies in their ability to learn from the past and leverage this historical knowledge, significantly improving and accelerating the optimization process in the current environment. 
Beyond tracking the moving optimum in DOPs, a subset of DOAs excels in identifying solutions that are robust against imminent environmental changes. 
These methods, commonly referred to as robust optimization over time (ROOT)~\cite{yazdani2023robust}, further utilize historical knowledge to predict the future quality and robustness of solutions, a necessary feature for DOPs where frequent modifications to the deployed solution are impractical or costly. 
Such challenges are especially prevalent in dynamic clustering for facility location problems, where frequently changing the locations of facilities (i.e., the cluster centers) is both undesirable and costly.

Meta-heuristic algorithms have established themselves as powerful tools for static clustering tasks, offering flexibility and robustness that are well-suited to finding optimal clustering in a fixed environment~\cite{hruschka2009survey}. 
Despite the large literature on using meta-heuristic algorithms for clustering in static environments and the fact that many real-world clustering scenarios inherently involve dynamic environments, the application of DOAs in clustering in dynamic environments remains largely unexplored. 
DOAs extend the capabilities of meta-heuristic algorithms by incorporating historical knowledge, which can be crucial for tackling the dynamic aspects of clustering tasks where data, structures, and even the number of clusters can change over time.
The limited application of DOAs in clustering in dynamic environments can be attributed to several factors. 
A primary challenge is the lack of benchmark datasets with the necessary diversity, controllable characteristics, and complexities to effectively test and evolve these algorithms. 
Unlike the static domain, which benefits from an abundance of datasets, dynamic clustering is yet to develop a comprehensive suite of benchmarks with the required flexibility and variety. 
Such benchmarks are essential for the progressive advancement of DOAs, enabling them to address a broader range of real-world challenges. 

Most existing DOAs are tailored for problems with simplistic assumptions, such as considering only one type of environmental change with uniform temporal and spatial severity and pattern~\cite{li2008GDBG,grefenstette1999evolvability,li2019open,branke1999memory,yazdani2021DOPsurveyPartB,yazdani2020benchmarking}. 
This creates a gap between academic research and the complex nature of practical applications, such as clustering in dynamic environments. 
To bridge this gap and unlock the full potential of DOAs for a wide range of clustering applications in dynamic environments, this paper introduces a new Dynamic Dataset Generator (DDG). 
This benchmark generator is designed to facilitate rigorous research and aid in the deployment of DOAs for clustering in dynamic environments, paving the way for these algorithms to meet the diverse and complex challenges presented by real-world dynamic settings.

DDG utilizes multiple dynamic Gaussian components (DGCs) for generating datasets. 
Each DGC is characterized by a wide array of controllable parameters, including its center location (mean),  standard deviation (width), weight, and rotation across all possible planes with specified angles. 
All these parameters as well as the number of DGCs and variables can change over time, influenced by various types of dynamics~\cite{yazdani2021DOPsurveyPartA,li2008GDBG}.

DDG distinguishes itself from previous dynamic benchmark generators~\cite{yazdani2021DOPsurveyPartB} in the field of DOPs and the dynamic synthetic dataset generators in the field of unsupervised learning~\cite{lu2018learning} under concept drift by its ability to simulate a wide range of dynamic scenarios. 
Traditional approaches often simulate simple and mostly homogeneous environmental changes, lacking the complexity observed in real-world dynamics~\cite{yazdani2021DOPsurveyPartA}. 
In contrast, DDG adeptly simulates a range of changes, from gradual, minor environmental changes to abrupt, significant transformations. 
This flexibility enables DDG to effectively mimic real-world dynamics, creating scenarios that are heterogeneous in both severity and frequency of changes. 

Furthermore, DDG advances beyond the conventional approach of fixed change frequency used in earlier benchmarks. Instead, it treats each change as probabilistic event, occurring with a specific likelihood at each time-tick (function evaluation), thus offering a more realistic representation of dynamic environments.

Additionally, DDG stands out as the first dynamic benchmark generator with the ability to control change correlation across all parameters of its DGCs. 
This distinctive capability enables DDG to simulate a wide array of dynamic scenarios, from those exhibiting rapid temporal severity to environments undergoing continuous change. 

DDG is a configurable benchmark generator, offering users the flexibility to customize problem instances by adjusting various parameters within the generator. 
It exhibits several key properties essential for an effective benchmark, as outlined in~\cite{bartz2020benchmarking}. 
These properties include the capability to generate a diverse range of problem instances with controllable dynamic characteristics and distributions (diversity) and controllable challenges and levels of difficulty (complexity variety). 
Furthermore, DDG simulates dynamic scenarios pertinent to various types of clustering in real-world dynamic environments, such as data with concept drift and facility location problems (relevance and representativity). 
Additionally, DDG is scalable in terms of the number of variables (scalability), and its source code is made publicly available at~\cite{yazdani2024DDGcodeMATLAB,yazdani2024DDGcodePython} (accessability).
Note, however, that one desirable property not encompassed by DDG is the availability of known optimal solutions. 
This is primarily due to the inherent nature of clustering problems as a domain of unsupervised learning: the absence of a known optimum solution or definitive ground truth for determining optimal clustering~\cite{hruschka2009survey}.
This necessitates a comparative analysis approach, focusing on the relative performance of algorithms when using DDG for benchmarking algorithms.

The remainder of this paper is organized as follows. 
Section~\ref{sec:RelatedWork} covers background information and related work.
Section~\ref{sec:proposedBenchmark} details the development and features of the proposed DDG. 
We discuss the performance measuring approach, settings, and preliminary findings in Section~\ref{sec:discussion}.
Finally, Section~\ref{sec:conclusion} concludes the paper.

\vspace{-12pt}
\section{Background and Related Work}
\label{sec:RelatedWork}

DOPs are prevalent in real-world scenarios, where evolving conditions often lead to a decline in the efficacy of previously deployed solutions~\cite{nguyen2011thesis}. 
This necessitates the identification and deployment of new solutions that are better suited to the current and upcoming states of the system. 
Despite the critical importance of real-world DOPs, a notable disparity exists between academic research and practical implementations in this field. 
This gap is largely attributed to the fact that algorithms designed within academic settings may not be fully equipped to confront the unique and complex challenges presented by practical DOPs.
The progression of DOAs for tackling DOPs is impeded by the limitations of existing dynamic benchmark problems. 
Most DOAs are tailored for these benchmarks, leaving their effectiveness in real-world scenarios, which often present more complex challenges, largely untested~\cite{yazdani2021DOPsurveyPartB}.

In the DOP field, dynamic benchmark generators based on moving peaks baseline functions are predominantly used for benchmarking DOAs~\cite{grefenstette1999evolvability,li2019open,branke1999memory}. 
These benchmark generators are preferred for their simplicity, configurability, and the ability to construct landscapes consisting of multiple peaks with changeable attributes like height, width, and location. 
However, a significant limitation of most benchmarks is their unrealistic landscape features, such as overly simplistic unimodal, symmetric, well-conditioned, and smooth peaks. 
Among the dynamic benchmark generators, the generalized moving peaks benchmark (GMPB)~\cite{yazdani2020benchmarking} has addressed these issue and is the only benchmark capable of controlling conditioning, symmetry, rotation, and ruggedness of its peaks.

However, despite advancements like those in GMPB, the dynamics used to simulate environmental changes in these benchmarks, still exhibit significant limitations. 
Real-world dynamic systems often involve a variety of environmental changes, each characterized by unique temporal and spatial severities, patterns, and areas of influence~\cite{yazdani2021DOPsurveyPartA}. 
This complexity is inadequately represented in most benchmarks, which tend to employ homogeneous changes. 
In the domain of ROOT, benchmarks do offer localized spatial severity for individual peaks, aiming to create a range of robustness~\cite{yazdani2023robust}. 
Yet, they fall short as the temporal severity remains uniform across all peaks and static over time, and spatial severity for each peak is also consistently homogeneous.

Another notable deficiency in current dynamic benchmark designs, such as those employing commonly used random dynamics, is their inability to effectively simulate scenarios with high temporal and low spatial severity. 
These scenarios, typical in continuously changing environments, are misrepresented as noise rather than meaningful environmental changes in current designs, resulting in a minimal cumulative effect over time due to frequent back-and-forth movements.

Furthermore, many existing baseline functions fail to adequately adapt the landscape when the number of peaks or dimensions is altered. 
For instance, adding or removing a dimension~\cite{li2008generalized} often leaves the optimum position in the existing dimensions unchanged, assuming no other types of changes occur. 
Similarly, the addition or removal of peaks tends to impact only locally~\cite{li2014adaptive}, without significantly affecting the broader landscape.

To address the limitations in existing benchmark problems that hinder the progression of DOAs, a direct transition into solving real-world problems with their complex challenges and indeterminate characteristics is not feasible. 
Instead, there is a pressing need for advanced benchmarks that offer more realistic landscapes and dynamics. 
Clustering in dynamic environments emerges as an ideal candidate for this purpose. 
It represents a subset of DOPs where clustering solutions must be continually updated to adapt to environmental changes. 
These problems cover a wide range of challenges and applications, from adapting to data exhibiting concept drift to optimizing real-time resource allocation in facility location problems. 
They call for sophisticated approaches capable of responding efficiently to ongoing changes, ensuring that clustering solutions stay relevant and effective in a dynamic setting.

In clustering datasets with concept drift, the underlying data distribution can change over time~\cite{suarez2023survey}.
Additionally, the number of clusters may increase due to emerging new clusters or decrease through the disappearance or merging of existing clusters. 
The amount and nature of noise and outliers, as well as rotations and the number of features, may also vary over time. 
Such changes can significantly alter the shape and size of clusters~\cite{silva2013data}.

Furthermore, clustering applications in problems like facility location problems in dynamic environments involve strategically relocating resources and facilities to efficiently cover an area and provide services~\cite{arabani2012facility}. 
Typically, the goal is to minimize the sum of distances between a set of objects, such as individuals, demands, and sensors,  and their closest facility, known as the $|\mathcal{P}|$-median in real-valued space~\cite{drezner1995facility}. 
After locating facilities, objects are assigned to their nearest facility, often using Euclidean distance or the shortest path. 
An example is strategic location and relocation of security forces in crowd monitoring and management, where facilities are often positioned to minimize response times to incidents, such as act of terrorist, in the crowd~\cite{martella2017current}.

In facility location problems, additional problem-specific objectives and constraints are usually considered, such as minimizing the maximum distance between a facility and an object or limiting the maximum number of objects in a cluster.
The dynamic nature of these clustering problems differs from clustering datasets with concept drift.
They are affected by extra time-varying parameters, such as environmental factors and priorities, and additional problem-specific objective functions and constraints that further influence the shape and size of clusters.
Moreover, the number of features in these problems is often fixed and typically corresponds to two, representing the geographical coordinates.

Using meta-heuristic approaches for clustering in static environments offers several advantages over traditional machine learning methods. 
These include independence from initial solutions, the flexibility to employ various clustering metrics as objective functions, and the capacity to satisfy multiple criteria simultaneously~\cite{handl2004evolutionary,mukhopadhyay2015survey}. 
Additionally, they allow for the integration of domain-specific knowledge into the objective function or constraints, further tailoring the solution to the problem at hand.
 In treating clustering as a general optimization problem, common metrics like the sum of intra-cluster distances~\cite{bandyopadhyay2002evolutionary}, the Davies-Bouldin index~\cite{davies1979cluster}, and the \(J_m\) criterion~\cite{bezdek2013pattern} are often used as objective functions. 
 These metrics, which typically frame clustering quality as a minimization problem, significantly influence the nature and methodology of the clustering process.

The methodology chosen for representing clustering solutions significantly influences the nature of the search space, which can be characterized by two key aspects. 
The first aspect relates to the type of data representation employed, which can be binary, integer, or continuous/real-valued~\cite{hruschka2009survey}. 
The second aspect involves the structural representation of the solution, falling into categories such as direct, prototype-based, or graph-based~\cite{handl2023evolutionary}. 
Each choice significantly influences the selection of suitable meta-heuristic algorithms and operators, as well as the effectiveness and complexity of the overall search process.

Meta-heuristic algorithms for clustering can also be categorized based on whether they assume a fixed number of clusters or dynamically determine the optimal number~\cite{handl2007evolutionary,garza2017improved}. 
Algorithms with a fixed number of clusters require prior knowledge or assumptions about the cluster count and are commonly used in scenarios where this information is known or can be estimated. 
In contrast, algorithms capable of determining the number of clusters are used when the number of clusters is not predetermined and needs to be optimized.

In the field of clustering in static environments, there is an abundance of datasets available for evaluating algorithms.
However, the landscape is markedly different for clustering in dynamic environments. 
Most benchmark datasets designed for learning under concept drift predominantly cater to classification problems, where  the labels may vary over time~\cite{gama2014survey}. 
Moreover, these datasets often fail to encompass a broad range of dynamic scenarios.
Firstly, the configurability of many synthetic datasets is limited. 
Parameters such as the number of variables and classes/clusters are typically fixed in many cases~\cite{lu2018learning}, though there are a few instances where they offer the flexibility to control the number of variables and classes/clusters~\cite{bifet2010moa}.
More significantly, the dynamics represented in these datasets tend to be restricted, focusing more on dataset-specific characteristics rather than systematic and controllable changes in underlying distributions. 

Dataset generators are also commonly used for creating data streams, with cores such as multivariate Gaussian distributions frequently employed for data generation~\cite{silva2013data}. 
Like dynamic benchmark problems, these generators often feature fixed temporal severity, where models are updated after generating a predefined number of data points~\cite{aggarwal2003framework}. 
In some generators, attributes such as the standard deviation and mean of Gaussian models are configured to change over time~\cite{aggarwal2003framework,wan2008weighted}. 
However, this evolution often fails to fully capture the complex dynamics of real-world environmental changes. These include heterogeneous alterations varying in spatial and temporal severity, as well as diverse domains of influence encompassing affected areas, the number of variables, and rotations.
In~\cite{aggarwal2004framework} the number of variables $d_G$ in the Gaussian models exceeds the dataset's variable count $d$. 
Data points are initially sampled using the larger variable set $d_G$, and subsequently, a subset of $d$ variables is selected for inclusion in the dataset. 
This methodology facilitates dynamic changes by periodically altering which dimensions of the Gaussian generators are included in the dataset.
Moreover, in~\cite{moulton2019clustering,webb2016characterizing}, a specific sampling method is utilized, involving the random generation of covariance matrices. 
For each Gaussian model, the Hellinger distance between the newly randomly generated covariance matrix and its predecessor is calculated. 
If this distance meets the predefined change severity, the covariance matrix is updated; otherwise, the sampling process is repeated until an appropriate matrix is identified. 
In this approach, the rejection-based covariance sampling becomes increasingly costly in higher dimensions.
Furthermore, it lacks a systematic approach to control the impacts resulting from updates to the covariance matrix.
Overall, current dataset generators designed for concept drift scenarios lack a systematic framework for controlling various dynamic aspects and parameters.

Therefore, there is a pressing need for a dataset generator framework capable of independently altering various aspects and parameters of the data generation process. 
This framework should enable handling multiple types of changes, with controllable temporal and spatial severities, parameters, areas of influence, and patterns of change. 
Additionally, controlled sampling within this framework ensures that changes in the dataset are also regulated. 
The framework must be adept at generating a wide array of changes, ranging from frequent gradual shifts to abrupt sudden alterations, encompassing local to global changes, and from incremental re-sampling data after gradual changes to completely re-sampling data from a drastically changed distribution. 
Crucially, all these variations should occur with the ability to control changes in every parameter of the generators and in every data point sampled.
In the next section, we introduce the Dynamic Dataset Generator (DDG), a tool specifically developed to meet the aforementioned needs and specifications.

\section{Proposed Dynamic Dataset Generator}
\label{sec:proposedBenchmark}

This section introduces the Dynamic Dataset Generator (DDG)\footnote{The Python and MATLAB source codes of DDG are publicly available and can be accessed at~\cite{yazdani2024DDGcodePython} and~\cite{yazdani2024DDGcodeMATLAB}, respectively.}, a novel tool tailored for the generation of dynamic datasets with controllable characteristics. 
Subsequent subsections provide detailed descriptions of DDG's key components: the data generator and dynamics.

\subsection{Data Generator}

DDG includes multiple Dynamic Gaussian Components (DGCs) for data generation. 
Each DGC in the $t$th environment is defined as follows:
\begin{align}
\label{eq:DGC}
    \mathcal{N}_i^{(t)}\left( \vec{c}_i^{(t)},\bm{\sigma}_{i}^{(t)},\bm\Theta^{(t)}_i \right) = \vec{r}\bm{\sigma}_{i}^{(t)}\mathbb{R}\left(\bm\Theta^{(t)}_i  \right)+\vec{c}_i^{(t)},
\end{align}
where $\mathcal{N}_i^{(t)}\left( \vec{c}_i^{(t)},\bm{\sigma}_{i}^{(t)},\bm\Theta^{(t)}_i \right)$ represents the $i$th DGC at time $t$. 
In this equation, $\vec{r}$ is a $d^{(t)}$-dimensional random vector, with each element drawn from a standard normal distribution $\mathcal{N}(0,1)$. 
The vector $\vec{c}_i^{(t)}$ denotes the center position (i.e., mean) of the $i$th DGC in the $t$th environment, and $d^{(t)}$ represents the number of variables at time $t$.
The vector $\bm\sigma_i^{(t)}$ is a $d^{(t)}$-dimensional vector representing the standard deviation, or the spread of the distribution, for the $i$th DGC across each dimension. 
The matrix $\bm\Theta_i^{(t)}$, a $d^{(t)} \times d^{(t)}$ matrix, plays a crucial role in defining the rotation of each component. 
All elements on and below the principal diagonal of $\bm\Theta_i^{(t)}$ are set to zero. 
Each off-diagonal element above the diagonal in the $j$th row and $k$th column (where $j < k$) specifies a rotation angle. 
This angle determines the degree of rotation for the projection of every point generated by the $i$th DGC in environment $t$, specifically onto the $x_j-x_k$ plane.

The matrix $\bm\Theta_i^{(t)}$ serves as an input to Algorithm~\ref{alg:RotationControlled}, denoted as $\mathbb{R}(\cdot)$ in Equation~\eqref{eq:DGC}. 
This algorithm generates a rotation matrix based on the specified angles in $\bm\Theta_i^{(t)}$. 
Readers interested in a more detailed explanation of this rotation matrix generation method are referred to~\cite{yazdani2023gnbg}.
In DDG, $m^{(t)}$ DGCs are used for data generation, each with a weight $w_i^{(t)}$ that controls the probability of generating a data point by the $i$th DGC $\left(p_i=w_{i}^{(t)}/\sum_{j=1}^{m^{(t)}} w_{j}^{(t)}\right)$.

\begin{algorithm}[!tp]
\footnotesize
$\mathbf{R}_{i}=\mathbf{I}_{d^{(t)} \times d^{(t)}}$\;
\For{$j=1$ to $d^{(t)}-1$}{\label{algline:Loop1}
\For{$k=j+1$ to $d^{(t)}$}{\label{algline:Loop2}
\If{$\bm\Theta_{i}^{(t)}(j,k)\neq 0$}{
$\mathbf{G}=\mathbf{I}_{d^{(t)} \times d^{(t)}}$\;
$\mathbf{G}(j,j)= \cos\left(\bm\Theta_{i}(j,k)\right)$\;
$\mathbf{G}(k,k) = \cos\left(\bm\Theta_{i}(j,k)\right)$\;
$\mathbf{G}(j,k) =  -\sin\left(\bm\Theta_{i}(j,k)\right)$\;
$\mathbf{G}(k,j) =  \sin\left(\bm\Theta_{i}(j,k)\right)$\;
$\mathbf{R}_{i}^{(t)} = \mathbf{R}_{i}^{(t)} \times \mathbf{G}$\;
}}}
Return $\mathbf{R}_{i}^{(t)}$\;
\caption{Generating rotation matrix $\mathbf{R}_{i}^{(t)}$\\
{\footnotesize\textbf{Input}: $d^{(t)}$ and $\bm\Theta_i^{(t)}$\\
\textbf{Output}: {$\mathbf{R}_{i}^{(t)}$}}
}
\label{alg:RotationControlled}
\end{algorithm}

\begin{figure}[t]
\centering
\begin{tabular}{cc}

    \subfigure[{\scriptsize Probability contour plot of a DDG with three DGCs. The warmer the color, the higher the probability; colder colors indicate lower probabilities.}]{\includegraphics[width=0.43\linewidth]{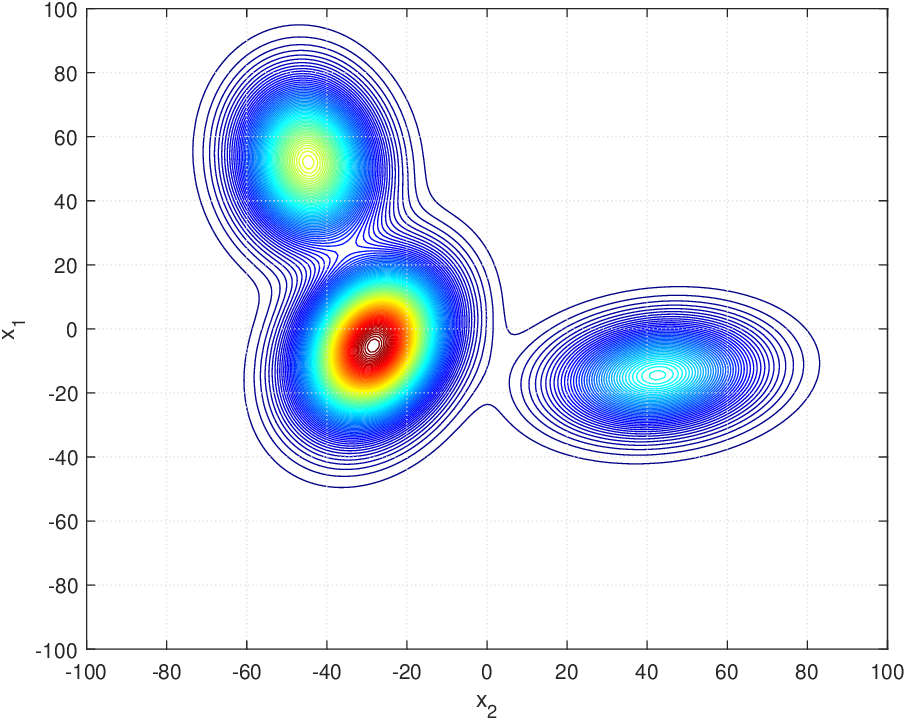}\label{fig:DDGcontour1}}
&
       \subfigure[{\scriptsize A dataset, including 1,000 samples, generated by the DDG shown in Figure~\ref{fig:DDGcontour1}.}]{\includegraphics[width=0.43\linewidth]{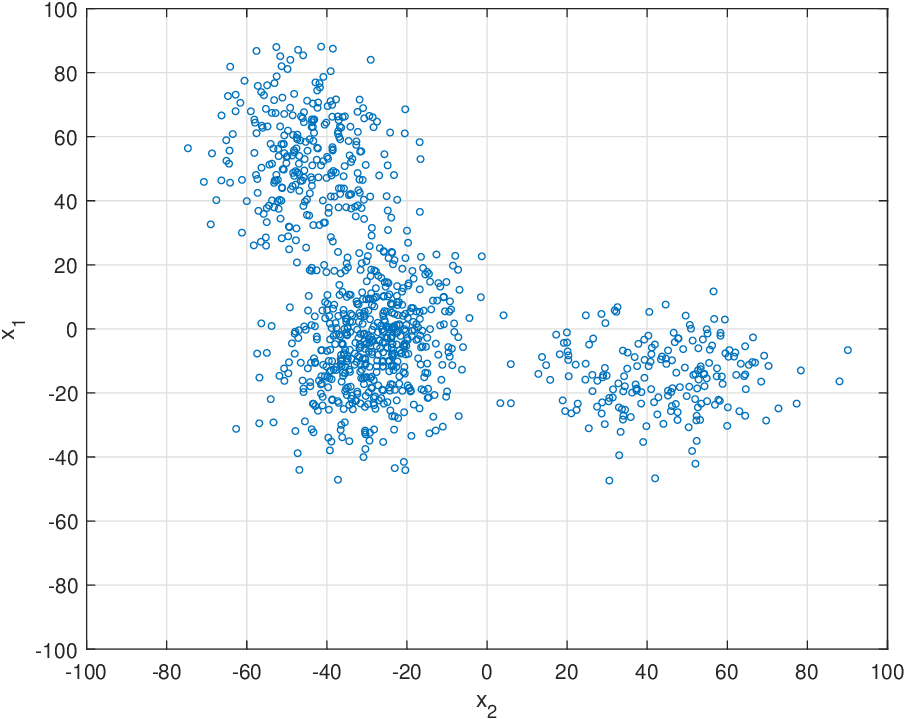}\label{fig:Sample1}}
\end{tabular}
\caption{Probability contour plot of a DDG with three DGCs, showcasing varied center positions, rotation angles, and weights. Each DGC has a $\bm\sigma$ set to [15,10]. 
The weights are distributed as [0.3, 0.5, 0.2] for the top, bottom left, and bottom right DGCs, respectively.
The dataset generated using this configuration is also displayed. 
}
\label{fig:DDGexample}
\end{figure}

\begin{figure}[t]
\centering
\begin{tabular}{cc}
    \subfigure[{\scriptsize A dataset, including 300 samples, generated by a DGC with $\vec{c}=[0,0]$ and $\bm\sigma=[20,20]$.}]{\includegraphics[width=0.4\linewidth]{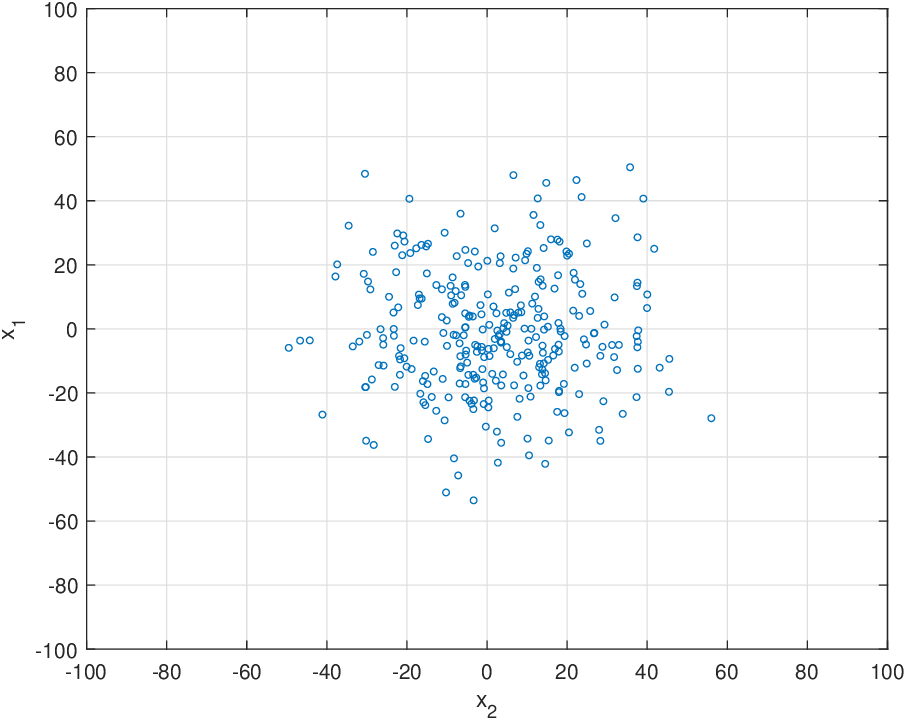}\label{fig:VariousSettingSample1}}
&
 \subfigure[{\scriptsize A dataset, including 300 samples, generated by a DGC with $\vec{c}=[-20,50]$ and $\bm\sigma=[7,7]$.}]{\includegraphics[width=0.4\linewidth]{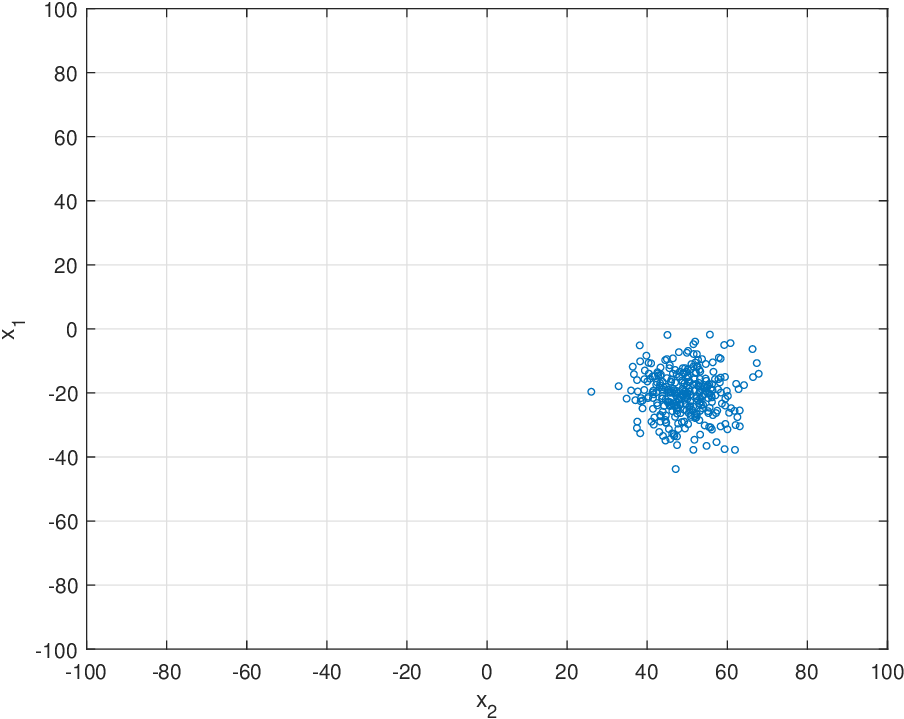}\label{fig:VariousSettingSample2}}
    \\
    \subfigure[{\scriptsize A dataset, including 300 samples, generated by a DGC with $\vec{c}=[0,0]$, $\bm\sigma=[7,20]$, and without rotation.}]{\includegraphics[width=0.4\linewidth]{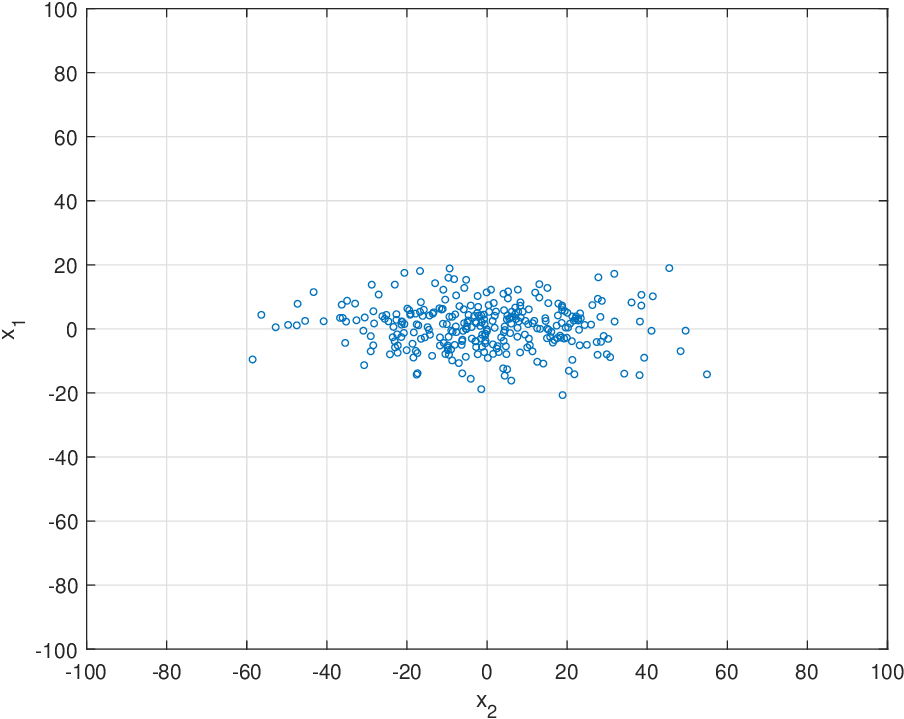}\label{fig:VariousSettingSample3}}
    &
    \subfigure[{\scriptsize A dataset, including 300 samples, generated by a DGC with $\vec{c}=[0,0]$, $\bm\sigma=[7,20]$, and rotated by $\frac{\pi}{4}$.}]{\includegraphics[width=0.4\linewidth]{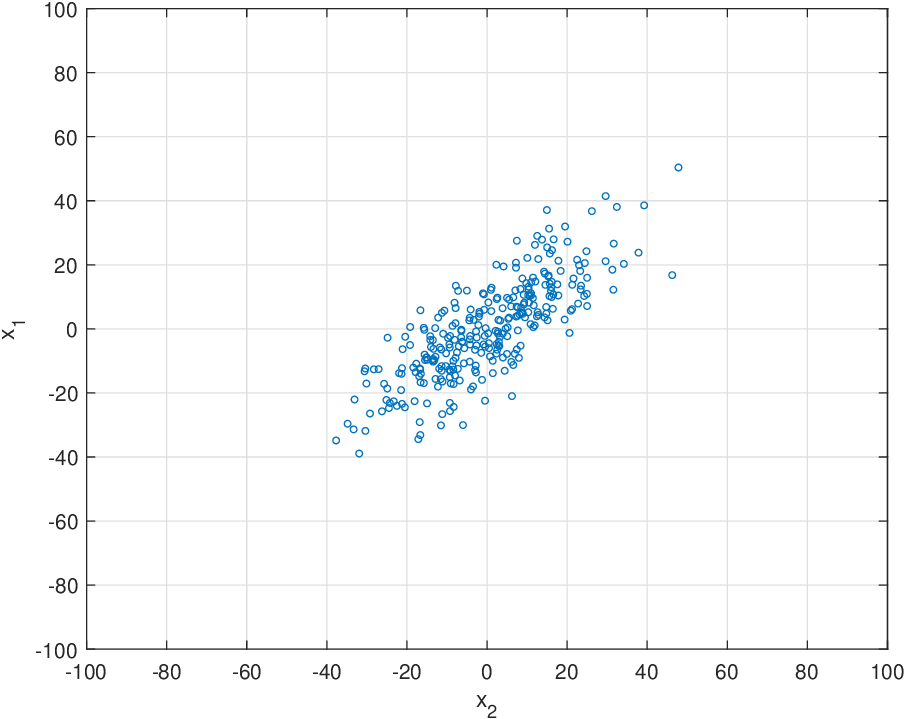}\label{fig:VariousSettingSample4}}
    \\
\end{tabular}
\caption{Impact of  various parameter settings on the distribution of 300 data generated by a DGC. 
}
\label{fig:DGCSettingsImpact}
\end{figure}

Figure~\ref{fig:DDGexample} presents an example of a DDG with three DGCs. 
In the DDG framework, the ensemble of DGCs collaboratively creates a landscape that emulates the characteristics of a multimodal joint probability distribution.
The influence of each DGC on this function is affected by its assigned weight, which modulates the corresponding Gaussian peak's volume by adjusting its height. 
This relationship is depicted in Figure~\ref{fig:DDGcontour1}, where, under the condition of identical $\bm\sigma_i$ values across DGCs, the peak heights are determined directly by their respective weight values. 
In the landscape formed by assembling DGCs, the probability assigned to any given point is the aggregate of the probabilities contributed by all DGCs at that point. 
Furthermore, Figure~\ref{fig:DGCSettingsImpact} showcases the impact of the parameters $\vec{c}_i$, $\bm\sigma_i$, and $\bm\Theta_i$ on shaping the data distribution produced by an individual DGC.

DDG's design with DGCs offers simplicity and flexibility in parameter manipulation to create various dynamic scenarios. 
Parameters such as the number of DGCs $(m^{(t)})$, probability weights for data generation ($w_i^{(t)}$), number of variables ($d^{(t)}$), DGCs' standard deviations ($\bm\sigma_i^{(t)}$), rotation angles ($\theta_{i,j,k}=\bm\Theta_i^{(t)}(j,k)$), and center positions ($\vec{c}_i^{(t)}$) can all dynamically change over time using diverse change patterns like random, chaotic, circular, or pendulum movements~\cite{yazdani2021DOPsurveyPartB}.

\subsubsection{Parameter Boundary Control in Dynamic Dataset Generator}

In DDG, each parameter has a predefined range. 
Although these ranges can also change over time, to maintain simplicity in DDG, we have chosen to keep them stationary.
However, they can be made dynamic with relative ease if required.

For each DGC in DDG, the center position in each dimension $j$ is bounded within $c_{i,j} \in [Lb_j, Ub_j]$, where $Lb_j$ and $Ub_j$ are the lower and upper bounds of the data values in the $j$th dimension, respectively. 
Each element of $\bm{\sigma}_i$ is constrained within $\sigma_{i,j} \in [\sigma_{\mathrm{min}}, \sigma_{\mathrm{max}}]$. 
Weight of DGCs is bounded within $w_{i} \in [w_{\mathrm{min}}, w_{\mathrm{max}}]$.
Each angle $\theta$ within $\bm{\Theta}_i$ is limited to $\theta \in [\theta_{\mathrm{min}}, \theta_{\mathrm{max}}]$.
In DDG, the number of variables, denoted as $d$, can also change over time, bounded within $d \in [d_{\mathrm{min}}, d_{\mathrm{max}}]$. 
Similarly, the number of DGCs $m$ can vary over time, bounded within $m \in [m_{\text{min}}, m_{\mathrm{max}}]$.

To ensure that the values of DDG's parameters remain within their specified bounds, a `reflect' method is employed~\cite{yazdani2020benchmarking}. 
This method operates as follows for a parameter $y$ within the range $[y_{\mathrm{min}}, y_{\mathrm{max}}]$:

\begin{equation}
y^{(t+1)} =
\begin{cases}
y^{(t+1)}, & \text{if } y^{(t+1)} \in [y_{\mathrm{min}}, y_{\mathrm{max}}] \\
2 \times y_{\mathrm{min}} - y^{(t+1)}, & \text{if } y^{(t+1)} < y_{\mathrm{min}} \\
2 \times y_{\mathrm{max}} - y^{(t+1)}, & \text{if } y^{(t+1)} > y_{\mathrm{max}}
\end{cases}.
\label{eq:BoundControl}
\end{equation}

\subsection{Simulating Dynamic Scenarios in Data Generation}

In this subsection, the simulation of dynamic scenarios in DDG is presented. 
The simulation encompasses three distinct aspects: 
\begin{inparaenum} 
\item gradual changes targeted at individual DGCs, 
\item changes with large impacts, and 
\item sampling schedules and data adjustment strategies.
\end{inparaenum}
Note that all the changes introduced in this section are categorized as random dynamics, where some of them are equipped with correlation factors that make them capable of generating small but frequent changes whose accumulated long term behavior is random. 
This mirrors the stochastic nature of most real-world dynamic systems~\cite{yazdani2021DOPsurveyPartA,yazdani2023robust}. 
However, these random dynamics in DDG can be substituted with other types of dynamics, such as re-appearing and chaotic dynamics to simulate a wider range of dynamic scenarios.

\subsubsection{Gradual Local Changes}
\label{sec:LocalChanges}

For each individual DGC $i$, several parameters can be dynamically adjusted: the DGC's location is modified by shifting $\vec{c}_i$, the DGC's width is varied through adjustments in $\bm\sigma_i$, weight is changed by modifying $w_i$ (works where $m\geq 2$), and the rotation is altered by updating $\bm\Theta_i$. 
Modifications in any of these DGC attributes lead to corresponding changes in the nature of the data sampled from that particular DGC.

For changing the center position of DGC $i$, a correlative dynamic approach is employed to relocate the center positions of promising regions over time. This is articulated in the following equations:
\begin{align}
\vec{c}_{i}^{~(t+1)}&=\vec{c}_{i}^{(t)} + n \tilde{s}_i  \vec{v}_{i}^{~(t+1)},\;\;\text{where}\label{eq:CenterDyanmic1} \\
\vec{v}_{i}^{~(t+1)}&=\frac{(1-\rho_i) \frac{\vec{r} }{\|\vec{r}\| }+ \rho_i  \vec{v}_{i}^{(t)}}{ \left\|(1-\rho_i) \frac{\vec{r} }{\|\vec{r}\| } + \rho_i \vec{v}_{i}^{(t)}\right\|}. \label{eq:correlativeOffset}
\end{align}
Here, $\tilde{s}_i$ represents the shift severity of DGC $i$, and $n \geq 0$ is a random number generated by a half-normal distribution $|\mathcal{N}(0,1)|$, $\vec{r}$ is a $d$-dimensional vector of random numbers from a normal distribution $\mathcal{N}(0,1)$,  $\|\cdot\|$ calculates the $l_2$-norm, the expression $\frac{\vec{r}}{\|\vec{r}\|}$ yields a $d$-dimensional unit vector with a random direction, and $\rho_i\in(0,1)$ is the correlation coefficient for DGC $i$. 
Note that based on Equation~\eqref{eq:correlativeOffset}, $\vec{v}_{i}$ is deliberately maintained as a unit vector. 
This design choice is intended to prevent the shift length from scaling with dimension, ensuring that the magnitude of change remains consistent regardless of the problem's dimensionality.

\begin{figure}[t]
\centering
\begin{tabular}{ccc}
   \subfigure[{\scriptsize $\rho=0.0$ }]{\includegraphics[width=0.3\linewidth]{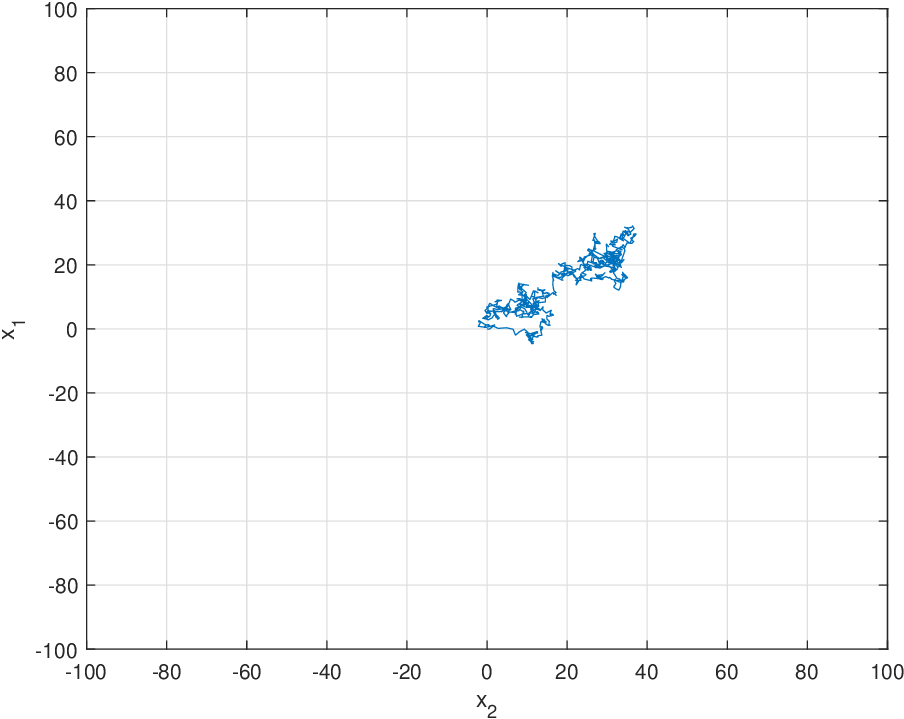}\label{fig:Lambda0-0}}
    &
    \subfigure[{\scriptsize $\rho=0.5$}]{\includegraphics[width=0.3\linewidth]{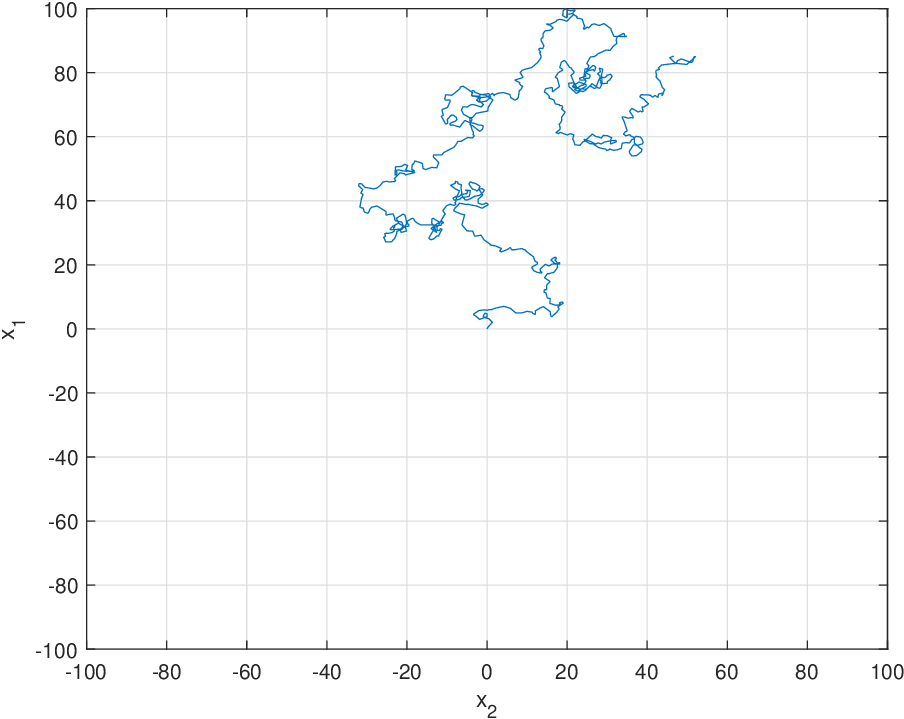}\label{fig:Lambda0-5}}
      &
    \subfigure[{\scriptsize $\rho=0.9$}]{\includegraphics[width=0.3\linewidth]{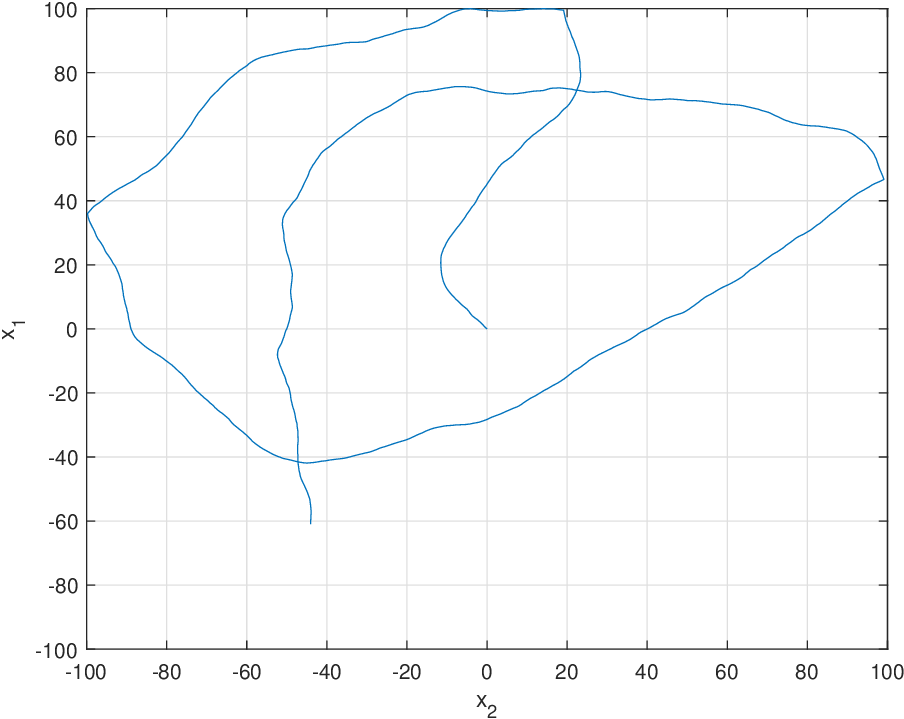}\label{fig:Lambda0-9}}
\end{tabular}
\caption{Trajectories of a DGC's center position, initially located at [0,0], over 1000 changes with a shift severity $\tilde{s}=1$, illustrated for three different values of the correlation coefficient $\rho$. 
}
\label{fig:peak_trajectories}
\vspace{-10pt}
\end{figure}

The use of correlative shifts is critical for simulating various environmental changes. 
Without such correlation, random offsets could lead to DGC centers oscillating chaotically.
Figure~\ref{fig:peak_trajectories} presents a comparative visualization of how the DGC's center position changes for different values of the correlation coefficient $\rho$.

To change the standard deviation value $\sigma_{i,j}^{(t)}$ within $\bm{\sigma}_i^{(t)}$, weight value $w_i$, and the angle $\theta_{i,j,k}^{(t)} = \bm{\Theta}_i^{(t)}(j,k)$, the following dynamics are employed:
\begin{align}
\sigma_{i,j}^{(t+1)} &= \sigma_{i,j}^{(t)} + \delta_{\sigma_{i,j}} n \tilde{\sigma}_i, \label{eq:SigmaDynamic}
\end{align}
\begin{align}
w_{i}^{(t+1)} &= w_{i}^{(t)} + \delta_{w_{i}} n \tilde{w}_i, \label{eq:HeightDynamic}
\end{align}
\begin{align}
\theta_{i,j,k}^{(t+1)} &= \theta_{i,j,k}^{(t)} + \delta_{\theta_{i,j,k}} n \tilde{\theta}_i.\label{eq:ThetaDynamic}
\end{align}
In the dynamic $y^{(t+1)} = y^{(t)} + \delta_{y} n \tilde{y}$ used in Equations~\eqref{eq:SigmaDynamic} to~\eqref{eq:ThetaDynamic}, $n \geq 0$ is a random number generated by a half-normal distribution $|\mathcal{N}(0,1)|$, $\delta_{y} \in \{-1,1\}$ is a direction factor for the parameter $y$ indicating whether $y^{(t+1)}$ is increasing or decreasing, $\tilde{\sigma}_i$,  $\tilde{w}_i$, and $\tilde{\theta}_i$ denote the severity of change for elements of $\bm{\sigma}_i$, weight, and angles in $\bm{\Theta}_i^{(t)}$, respectively. 
At each time-tic (function evaluation), there is a probability $p_{l,i}$ of inverting the direction factors $\delta_{w_{i}}$, $\delta_{\sigma_{i,j}}$, and $\delta_{\theta_{i,j,k}}$.
Additionally, $\delta_{y}$ is inverted when the updated value exceeds its boundary and is reflected back within its range using Equation~\eqref{eq:BoundControl}. 
A $p_{y}$ value of zero implies that the direction of change for parameter $y$ remains constant until it reaches a boundary, whereas a value of 0.5 shows a uniformly random direction at each change. 
Essentially, $p_{y}$ determines the correlation between successive changes in the value of parameter $y$.
The changes outlined in Equations~\eqref{eq:CenterDyanmic1}, \eqref{eq:SigmaDynamic}, \eqref{eq:HeightDynamic}, and \eqref{eq:ThetaDynamic} are applied to DGC $i$ with a certain probability $\mathfrak{p}_{l,i}$ at each function evaluation.
This probability dictates the likelihood of these changes being applied to each specific DGC.

\begin{figure}[t]
\centering
\begin{tabular}{cc}
   \subfigure[{\scriptsize  $\tilde{y} = 1$  and $p_y=0$}]{\includegraphics[width=0.47\linewidth]{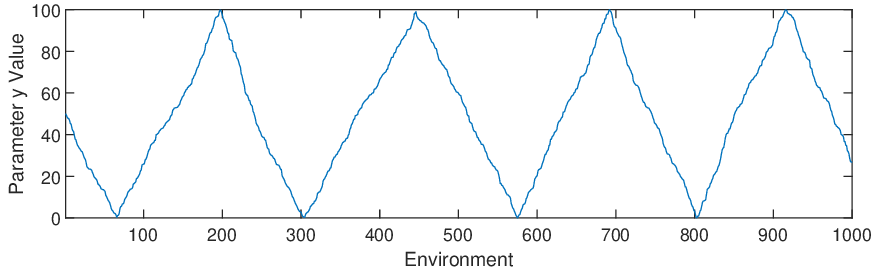}\label{fig:p0-0}}
    &
    \subfigure[{\scriptsize $\tilde{y} = 1$  and $p_y=0.1$}]{\includegraphics[width=0.47\linewidth]{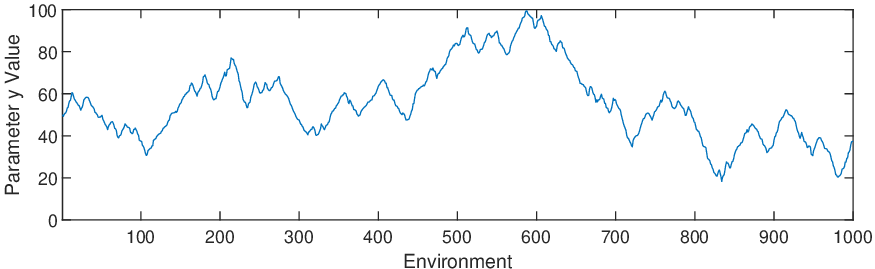}\label{fig:p0-1}}
    \\
      \subfigure[{\scriptsize $\tilde{y} = 1$  and $p_y=0.5$}]{\includegraphics[width=0.47\linewidth]{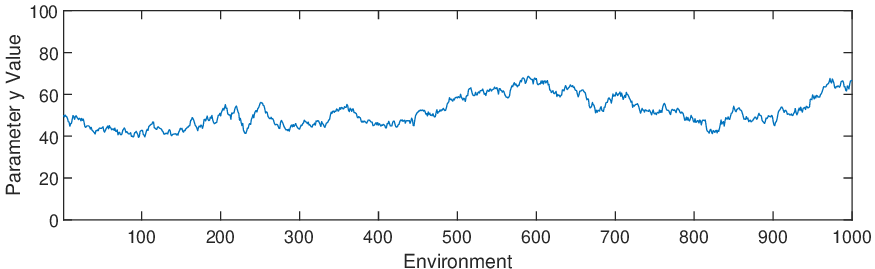}\label{fig:p0-5s1}}
    &
    \subfigure[{\scriptsize $\tilde{y} = 0.1$  and $p_y=0.5$}]{\includegraphics[width=0.47\linewidth]{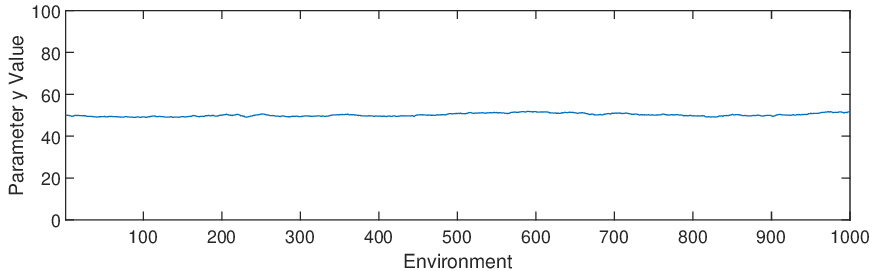}\label{fig:p0-5s0-1}}
\end{tabular}
\caption{Dynamic changes of a parameter $y$ where $y\in[0,100]$ and $y^{(1)}=50$, using the dynamic used in Equations~\eqref{eq:SigmaDynamic} to~\eqref{eq:ThetaDynamic} over 1000 changes with various settings of change severity $\tilde{y}$ and the probability $p_y$ for altering the direction factor $\delta_y$. 
}
\label{fig:Dynamics4Peaks}
\vspace{-10pt}
\end{figure}

Figure~\ref{fig:Dynamics4Peaks} illustrates the behavior of the dynamics presented in Equations~\eqref{eq:SigmaDynamic} to~\eqref{eq:ThetaDynamic}, particularly how they respond to variations in the probability of altering the direction factor.
This probability factor serves a role analogous to $\rho$ in Equation~\eqref{eq:correlativeOffset}, as both parameters influence the correlation of changes over time. 
Note that when the correlation is minimal (representing a fully random change) and the change severity is low, the parameters tend to fluctuate slightly around their initial values without significant cumulative change, as shown in Figure~\ref{fig:p0-5s0-1}. 
This observation highlights the importance of employing more correlated changes in parameters to simulate scenarios with frequent yet gradual changes, a characteristic common in many real-world problems. 

\subsubsection{Changes With Large Impacts}
\label{sec:LargeChanges}

The previous section focused on local, correlation-based changes, each specifically targeting individual DGCs with variable severity. 
In contrast, this section introduces changes with significant impact. 
This includes alterations affecting all DGCs with large severity, variations in the number of variables and DGCs, as well as changes in the number of clusters. 

To simulate abrupt global changes with significant severity across all DGCs, DDG employs the following dynamics to modify the attributes of all DGCs:
\begin{align}
\vec{c}_{i}^{(t+1)}&=\vec{c}_{i}^{(t)} + (2\mathcal{B}(\alpha,\beta)-1)\widehat{s}\frac{ \vec{r}}{\|\vec{r} \|}, \label{eq:Gcenter}
\end{align}
\begin{align}
w_{i}^{(t +1)}&=w_{i}^{(t)}  + (2\mathcal{B}(\alpha,\beta)-1)\widehat{w}, \label{eq:Gheight} 
\end{align}
\begin{align}
\sigma_{i,j}^{(t +1)}&=\sigma_{i,j}^{(t)}  +  (2\mathcal{B}(\alpha,\beta)-1)\widehat{\sigma}, \label{eq:Gwidth} 
\end{align}
\begin{align}
\theta_{i,j,k}^{(t +1)}&= \theta_{i,j,k}^{(t)}+(2\mathcal{B}(\alpha,\beta)-1)\widehat{\theta} .\label{eq:Gangle}
\end{align}
Here, $2\mathcal{B}(\alpha,\beta)-1$ symbolizes a Beta distribution, generating random values within the range of $[-1,1]$. For these dynamics, small and equal values for $\alpha$ and $\beta$ are selected to generate a heavy-tail and symmetric distribution, typically setting $\alpha=\beta<0.2$. 
This choice is instrumental in creating significant changes, as the heavy-tail distribution emphasizes more extreme values, reducing the likelihood of negligible changes due to random values near zero.
The vector $\vec{r}$ is comprised of random numbers generated by $\mathcal{N}(0,1)$.
The \emph{global} severity values for weight, width, shift, and angle of all DGCs are denoted by $\widehat{w}$, $\widehat{\sigma}$, $\widehat{s}$, $\widehat{\theta}$, respectively.
These global change severity parameters are set to be substantially larger than those used for the local, correlation-based changes detailed in Section~\ref{sec:LocalChanges}.
The set of changes defined in Equations~\eqref{eq:Gcenter} to~\eqref{eq:Gangle} are applied to all DGCs with a specific likelihood, denoted as $\mathfrak{p}_g$. 

Alterations in the number of DGCs not only add or remove specific groups of data, impacting the data distribution locally, but also globally affect the dataset by altering the normalized weights of all existing DGCs.
In DDG, the following dynamic is used for changing the number of DGCs $m$ over time: 
\begin{align}
m^{(t+1)}=m^{(t)} + b\tilde{m}, \label{eq:ChangePeakNumber}
\end{align}
where $b\in\{-1,1\}$ is a binary variable randomly selected with equal probability to be either -1 or 1, and $\tilde{m}$, a positive integer, determines the number of DGCs to be added or removed. 
This integer can be randomly selected from a set of predefined values, although a common choice is 1. 
The frequency of changes by Equation~\eqref{eq:ChangePeakNumber} is controlled by the probability $\mathfrak{p}_m$, which dictates the likelihood of a change in the number of DGCs occurring at each function evaluation.
If the number of DGCs $m\in \{m_\mathrm{min},\cdots,m_\mathrm{max}\}$ is reduced, a randomly selected DGC is removed.
Conversely, when introducing a new DGC, its parameters are randomly initialized.
Note that changes in the number of DGCs do not directly correlate with the optimal number of clusters for the clustering algorithm that also optimize the number of clusters. 
The reason lies in the fact that the optimal cluster count does not necessarily align with the number of DGCs in the dataset generation. 
This discrepancy arises from various factors, such as the potential merging or overlapping of clusters, or the overall data distribution suggesting a different clustering arrangement. 
Furthermore, differing clustering metrics, each with its distinct definition of a ``cluster,'' can result in varied interpretations of the optimal cluster count.

Changes in the number of variables, particularly relevant in data with concept drift, can significantly alter the clustering landscape. 
This dynamic is not common in facility location clustering, which typically involves 2-dimensional geographical locations. 
To adjust the number of variables, $d$, within the range of $\{d_\mathrm{min},\cdots,d_\mathrm{max}\}$, we employ a mechanism similar to Equation~\eqref{eq:ChangePeakNumber}:
\begin{align}
d^{(t+1)}=d^{(t)} + b\tilde{d}. \label{eq:ChangeVariable}
\end{align}
The probability $\mathfrak{p}_d$ determines the likelihood of a change in the number of variables at each function evaluation, with $\tilde{d}$ typically set to 1. 
When reducing the number of variables, a random variable $j$ is selected for removal, eliminating the corresponding $j$th elements from $\vec{c}$, $\bm\sigma$, and both the $j$th column and row in $\bm\Theta$ for all DGCs. 
Conversely, introducing a new variable involves randomly initializing it as the $j$th variable, followed by the addition of corresponding randomized values to $\vec{c}$, $\bm\sigma$, and the $j$th column and row of $\bm\Theta$ for all DGCs. 
In this process, indices $k\geq j$ are incremented by 1. 

Additionally, in certain scenarios, the number of clusters is an external parameter of the dataset and can change over time, influencing the problem space for clustering algorithms. 
While in some clustering scenarios, the number of clusters is an optimization target, in others, it is determined by external factors, such as decisions made by authorities or the number of available facilities.
To model these scenarios in DDG, the number of clusters, denoted as $\kappa$ within the range $\{\kappa_\mathrm{min},\cdots,\kappa_\mathrm{max}\}$, can be adjusted dynamically. 
The change is governed by the equation:
\begin{align}
\kappa^{(t+1)}=\kappa^{(t)} + b\tilde{\kappa}. \label{eq:ChangeClusterNum}
\end{align}
For changing the number of clusters, a parameter $\mathfrak{p}_\kappa$ is used as the probability of a change occurring at each function evaluation. 
The parameter $\tilde{\kappa}$, typically set to 1, determines the magnitude of change. 

\subsubsection{Synchronization and Management of Data Generation}
\label{sec:SmplingMnagement}

In the DDG framework, various strategies are employed to reflect the changes in DGCs to the dataset $\mathbf{D}$. While the size of $\mathbf{D}$ can be either fixed or variable over time, for simplicity, the initial version of DDG maintains a fixed dataset size. 
Data generation relies on the current state of the DGCs as described in Equation~\eqref{eq:DGC}. 
Each strategy for updating $\mathbf{D}$ is designed for specific simulation objectives and set of dynamics used.

Data generation utilizes the current state of DGCs in Equation~\eqref{eq:DGC}. 
To generate a data point, DGCs' weights are normalized by dividing them by the sum $\sum_{j=1}^{m^{(t)}} w_{j}^{(t)}$, forming the basis for probabilistically selecting a DGC for data generation. 

To incorporate gradual changes (Section~\ref{sec:LocalChanges}) into the data, generation is probabilistically triggered at each time tick, based on a predetermined likelihood $\mathfrak{p}_s$. 
In scenarios where new data continuously arrives, the number of new data points sampled is a fraction $\mathfrak{n}\%$ of $\mathbf{D}$'s size. 
Upon adding new data, an equal proportion of the oldest data is removed, following a First-In-First-Out approach. 
The value of $\mathfrak{n}\%$ should be chosen in relation to $\mathfrak{p}_s$, with smaller $\mathfrak{n}\%$ for more frequent sampling at larger $\mathfrak{p}_s$, and vice versa.

Significant changes (Section~\ref{sec:LargeChanges}) trigger a comprehensive data generation process. 
Post major changes, as triggered by Equations~\eqref{eq:Gcenter} to~\eqref{eq:ChangeVariable}, the entire dataset $\mathbf{D}$ is re-sampled to align with the new environmental state.

\section{Discussion: Performance Measures, Settings, and Preliminary Findings}
\label{sec:discussion}

The DDG framework has been carefully developed to benchmark algorithms designed for clustering in dynamic environments. 
It encompasses a comprehensive array of configurable parameters, facilitating the simulation of diverse dynamic scenarios tailored to specific research objectives. 
Table~\ref{tab:DDGparameters} in the appendix summarizes all parameters of DDG, and provides suggested values and ranges for these parameters.

In the DDG framework, the objective functions and methodologies for representing clustering solutions align with those employed in static environment clustering. 
However, a distinct approach is required for performance measurement. 
Given DDG's reliance on multiple event-based changes, each with varying likelihoods, traditional performance indicators that rely on single values per environment, such as average fitness of the best found solution in each environment $t$~\cite{trojanowski1999searching}, become less suitable as the length of environments varies over time. 
Therefore, we need to focus on performance indicators that measure performance over all function evaluations, such as offline performance~\cite{branke1999memory}.
In addition, a challenge in clustering problems as unsupervised learning is the absence of a known optimum solution or definitive ground truth for determining optimal clustering~\cite{hruschka2009survey}. 
Therefore, instead of measuring the deviation from optimum performance, such as error-based metrics, the focus must be on comparing algorithmic performance against each other.
Furthermore, in the context of ROOT, even in optimization problems where the optimal solutions are known, the optimal robustness remains undefined~\cite{yazdani2023robust}. 
Another challenge in DOPs is the variability in performance measures due to stochastic events and randomness in problem generation, which can lead to unstable results and potentially unfair comparisons. 
To address this, DDG employs a consistent random stream in each run to ensure fairness and reliability in the comparisons.

In dynamic scenarios, where the goal is to find the best clustering solution for each environmental state, offline performance-based metrics~\cite{yazdani2021DOPsurveyPartB}, such as the average of the best sum of intra-cluster distances over all function evaluations, are employed.
Following any change in the dataset $\mathbf{D}$, the current best solution undergoes re-evaluation.
After each clustering solution evaluation, the best performing solution is updated if the newly evaluated solution surpasses it.
For robust clustering over time (i.e., ROOT in clustering), we focus on the deployed solution where the performance evaluation relies on the average number of function evaluations during which the deployed solution maintains acceptable quality. 

Utilizing DDG with the default settings outlined in Table~\ref{tab:DDGparameters} in the appendix, we conducted a preliminary study to investigate the effectiveness of several DOAs in addressing the challenges of clustering in dynamic environments. 
Ten recent DOAs from the EDOLAB platform~\cite{peng2023evolutionary} were tested, focusing on real-valued space cluster center positioning and the sum of intra-cluster distances as the objective function. 
Our findings highlighted significant shortcomings in the current DOAs, particularly in continuously changing environments. 
The inherently reaction-based nature of these algorithms~\cite{nguyen2012evolutionary,yazdani2021DOPsurveyPartA}, and the frequent activation of their change-reaction mechanisms, proved counterproductive.

The detection of different types of environmental changes emerged as another significant challenge. 
Common re-evaluation-based change detection methods~\cite{richter2009detecting} are inadequate in continuously changing scenarios. 
Moreover, scenarios with multiple heterogeneous changes pose a heightened challenge, demanding specific detection and response strategies for each type of change. 
For instance, algorithms that also optimize the number of clusters~\cite{mukhopadhyay2015survey} struggled with identifying changes in the optimal cluster count.

Moreover, many algorithms are limited as they are developed with the assumption of a singular type of environmental change~\cite{yazdani2021DOPsurveyPartB}. 
This limitation renders the use of historical knowledge either ineffective or detrimental to the optimization process in scenarios characterized by multiple heterogeneous dynamics. 
Additionally, global diversity control mechanisms, such as the widely employed exclusion mechanisms~\cite{blackwell2006multiswarm}, are crucial in many DOAs for preserving diversity and optimizing computational resource usage. 
However, these mechanisms fail to differentiate between similar clustering solutions that vary merely in the permutation of cluster centers.

As the data distribution changes over time, the effective search range for DOAs also needs to adapt accordingly to align with the current data distribution. 
Failure to do so can result in the generation of low-quality individuals or sub-populations, especially after employing commonly used randomization methods for enhancing global diversity~\cite{blackwell2008particle}. 
Additionally, the optimal values of many DOA parameters are often contingent upon the search range.
Therefore, it becomes necessary to update these parameters over time.  
Furthermore, challenges arise when utilizing commonly used mechanisms in DOAs that are based on Euclidean distances, such as convergence detection~\cite{yazdani2023species}. 
However, in scenarios where the search space is a hyper-rectangle with significant disparity in different dimensions, the effectiveness of Euclidean distance-based mechanisms deteriorates.

Considering the multitude of challenges presented by DDG, it becomes evident that developing a single, comprehensive solution to simultaneously address all these challenges is not currently feasible. 
A more strategic approach involves tackling each challenge individually, creating specific mechanisms and components tailored to address each one effectively. 
In line with this, leveraging DDG's high configurability is crucial for generating instances with isolated, specific challenges. 
To disable any type of dynamic in DDG, the likelihood parameter for that dynamic can be set to zero. Additionally, to prevent changes to a specific parameter, its change severity value can be set to zero.

\section{Conclusion}
\label{sec:conclusion}

In this paper, we introduced the Dynamic Dataset Generator (DDG), a comprehensive benchmark generator tool crafted for creating dynamic datasets with known and controllable characteristics, specifically tailored for the evaluation of algorithms in clustering in dynamic environments. 
At the heart of DDG are multiple Dynamic Gaussian Components (DGCs), which facilitate a broad range of dynamic scenarios. 
These range from subtle, continuous environmental changes to abrupt, substantial transformations.
DDG is uniquely equipped to generate dynamic scenarios characterized by heterogeneous changes. 
These changes encompass a variety of temporal and spatial severities, patterns, and areas of influence, effectively capturing both local and global dynamics. 
Through this diverse and flexible approach, DDG serves as a vital tool for realistically simulating and assessing the performance of clustering algorithms in dynamic, real-world-like scenarios.

The next step involves developing a comprehensive suite of preset dynamic scenarios with DDG, aimed at standardizing research directions and facilitating ease of use for broader applications in the field.
Additionally, a future work includes expanding DDG for generating dynamic classification and regression problems.

\bibliography{bib}

\begin{thebibliography}{10}
\providecommand{\url}[1]{#1}
\csname url@samestyle\endcsname
\providecommand{\newblock}{\relax}
\providecommand{\bibinfo}[2]{#2}
\providecommand{\BIBentrySTDinterwordspacing}{\spaceskip=0pt\relax}
\providecommand{\BIBentryALTinterwordstretchfactor}{4}
\providecommand{\BIBentryALTinterwordspacing}{\spaceskip=\fontdimen2\font plus
\BIBentryALTinterwordstretchfactor\fontdimen3\font minus
  \fontdimen4\font\relax}
\providecommand{\BIBforeignlanguage}[2]{{%
\expandafter\ifx\csname l@#1\endcsname\relax
\typeout{** WARNING: IEEEtran.bst: No hyphenation pattern has been}%
\typeout{** loaded for the language `#1'. Using the pattern for}%
\typeout{** the default language instead.}%
\else
\language=\csname l@#1\endcsname
\fi
#2}}
\providecommand{\BIBdecl}{\relax}
\BIBdecl

\bibitem{jain1999data}
A.~K. Jain, M.~N. Murty, and P.~J. Flynn, ``Data clustering: a review,''
  \emph{ACM Computing Surveys}, vol.~31, no.~3, pp. 264--323, 1999.

\bibitem{xu2005survey}
R.~Xu and D.~Wunsch, ``Survey of clustering algorithms,'' \emph{IEEE
  Transactions on neural networks}, vol.~16, no.~3, pp. 645--678, 2005.

\bibitem{moulton2019clustering}
R.~H. Moulton, H.~L. Viktor, N.~Japkowicz, and J.~Gama, ``Clustering in the
  presence of concept drift,'' in \emph{Machine Learning and Knowledge
  Discovery in Databases}.\hskip 1em plus 0.5em minus 0.4em\relax Springer,
  2019, pp. 339--355.

\bibitem{zhan2020online}
X.~Zhan, J.~Xie, Z.~Liu, Y.-S. Ong, and C.~C. Loy, ``Online deep clustering for
  unsupervised representation learning,'' in \emph{Proceedings of the IEEE/CVF
  conference on computer vision and pattern recognition}, 2020, pp. 6688--6697.

\bibitem{karatas2021dynamic}
M.~Karatas, ``A dynamic multi-objective location-allocation model for search
  and rescue assets,'' \emph{European Journal of Operational Research}, vol.
  288, no.~2, pp. 620--633, 2021.

\bibitem{li2022evolutionary}
T.~Li, L.~Chen, C.~S. Jensen, T.~B. Pedersen, Y.~Gao, and J.~Hu, ``Evolutionary
  clustering of moving objects,'' in \emph{International Conference on Data
  Engineering}.\hskip 1em plus 0.5em minus 0.4em\relax IEEE, 2022, pp.
  2399--2411.

\bibitem{handl2007ant}
J.~Handl and B.~Meyer, ``Ant-based and swarm-based clustering,'' \emph{Swarm
  Intelligence}, vol.~1, pp. 95--113, 2007.

\bibitem{nguyen2011thesis}
T.~T. Nguyen, ``Continuous dynamic optimisation using evolutionary
  algorithms,'' Ph.D. dissertation, University of Birmingham, 2011.

\bibitem{yazdani2021DOPsurveyPartA}
D.~Yazdani, R.~Cheng, D.~Yazdani, J.~Branke, Y.~Jin, and X.~Yao, ``A survey of
  evolutionary continuous dynamic optimization over two decades -- part {A},''
  \emph{IEEE Transactions on Evolutionary Computation}, vol.~25, no.~4, pp.
  609--629, 2021.

\bibitem{yazdani2018thesis}
D.~Yazdani, ``Particle swarm optimization for dynamically changing environments
  with particular focus on scalability and switching cost,'' Ph.D.
  dissertation, Liverpool John Moores University, Liverpool, UK, 2018.

\bibitem{yazdani2023robust}
D.~Yazdani, M.~N. Omidvar, D.~Yazdani, J.~Branke, T.~T. Nguyen, A.~H. Gandomi,
  Y.~Jin, and X.~Yao, ``Robust optimization over time: A critical review,''
  \emph{IEEE Transactions on Evolutionary Computation}, Early access, 2023.

\bibitem{hruschka2009survey}
E.~R. Hruschka, R.~J. Campello, A.~A. Freitas \emph{et~al.}, ``A survey of
  evolutionary algorithms for clustering,'' \emph{IEEE Transactions on systems,
  man, and cybernetics, Part C}, vol.~39, no.~2, pp. 133--155, 2009.

\bibitem{li2008GDBG}
C.~Li, S.~Yang, T.~T. Nguyen, E.~L. Yu, X.~Yao, Y.~Jin, H.-G. Beyer, and P.~N.
  Suganthan, ``Benchmark generator for cec'2009 competition on dynamic
  optimization,'' Center for Computational Intelligence, Tech. Rep., 2008.

\bibitem{grefenstette1999evolvability}
J.~J. Grefenstette, ``Evolvability in dynamic fitness landscapes: a genetic
  algorithm approach,'' in \emph{Congress on Evolutionary Computation},
  vol.~3.\hskip 1em plus 0.5em minus 0.4em\relax IEEE, 1999, pp. 2031--2038.

\bibitem{li2019open}
C.~Li, T.~T. Nguyen, S.~Zeng, M.~Yang, and M.~Wu, ``An open framework for
  constructing continuous optimization problems,'' \emph{IEEE Transactions on
  Cybernetics}, vol.~49, no.~6, pp. 2316--2330, 2018.

\bibitem{branke1999memory}
J.~Branke, ``Memory enhanced evolutionary algorithms for changing optimization
  problems,'' in \emph{Congress on Evolutionary Computation}, vol.~3.\hskip 1em
  plus 0.5em minus 0.4em\relax IEEE, 1999, pp. 1875--1882.

\bibitem{yazdani2021DOPsurveyPartB}
D.~Yazdani, R.~Cheng, D.~Yazdani, J.~Branke, Y.~Jin, and X.~Yao, ``A survey of
  evolutionary continuous dynamic optimization over two decades -- part {B},''
  \emph{IEEE Transactions on Evolutionary Computation}, vol.~25, no.~4, pp.
  630--650, 2021.

\bibitem{yazdani2020benchmarking}
D.~Yazdani, M.~N. Omidvar, R.~Cheng, J.~Branke, T.~T. Nguyen, and X.~Yao,
  ``Benchmarking continuous dynamic optimization: survey and generalized test
  suite,'' \emph{IEEE Transactions on Cybernetics}, vol.~52, no.~5, pp.
  3380--3393, 2022.

\bibitem{lu2018learning}
J.~Lu, A.~Liu, F.~Dong, F.~Gu, J.~Gama, and G.~Zhang, ``Learning under concept
  drift: A review,'' \emph{IEEE Transactions on Knowledge and Data
  Engineering}, vol.~31, no.~12, pp. 2346--2363, 2018.

\bibitem{bartz2020benchmarking}
T.~Bartz-Beielstein, C.~Doerr, D.~v.~d. Berg, J.~Bossek, S.~Chandrasekaran,
  T.~Eftimov, A.~Fischbach, P.~Kerschke, W.~La~Cava, M.~Lopez-Ibanez
  \emph{et~al.}, ``Benchmarking in optimization: Best practice and open
  issues,'' \emph{arXiv preprint arXiv:2007.03488}, 2020.

\bibitem{yazdani2024DDGcodeMATLAB}
\BIBentryALTinterwordspacing
D.~Yazdani, ``The matlab source code of dynamic dataset generator,'' in
  \emph{GIThub}, 2024. [Online]. Available:
  \url{https://github.com/Danial-Yazdani/DDG-MATLAB}
\BIBentrySTDinterwordspacing

\bibitem{yazdani2024DDGcodePython}
\BIBentryALTinterwordspacing
------, ``The python source code of dynamic dataset generator,'' in
  \emph{GIThub}, 2024. [Online]. Available:
  \url{https://github.com/Danial-Yazdani/DDG-Python}
\BIBentrySTDinterwordspacing

\bibitem{li2008generalized}
C.~Li and S.~Yang, ``A generalized approach to construct benchmark problems for
  dynamic optimization,'' in \emph{Simulated Evolution and Learning}.\hskip 1em
  plus 0.5em minus 0.4em\relax Springer, 2008, pp. 391--400.

\bibitem{li2014adaptive}
C.~Li, S.~Yang, and M.~Yang, ``An adaptive multi-swarm optimizer for dynamic
  optimization problems,'' \emph{Evolutionary Computation}, vol.~22, no.~4, pp.
  559--594, 2014.

\bibitem{suarez2023survey}
A.~L. Su{\'a}rez-Cetrulo, D.~Quintana, and A.~Cervantes, ``A survey on machine
  learning for recurring concept drifting data streams,'' \emph{Expert Systems
  with Applications}, vol. 213, p. 118934, 2023.

\bibitem{silva2013data}
J.~A. Silva, E.~R. Faria, R.~C. Barros, E.~R. Hruschka, A.~C.~d. Carvalho, and
  J.~Gama, ``Data stream clustering: A survey,'' \emph{ACM Computing Surveys},
  vol.~46, no.~1, pp. 1--31, 2013.

\bibitem{arabani2012facility}
A.~B. Arabani and R.~Z. Farahani, ``Facility location dynamics: An overview of
  classifications and applications,'' \emph{Computers \& Industrial
  Engineering}, vol.~62, no.~1, pp. 408--420, 2012.

\bibitem{drezner1995facility}
Z.~Drezner, \emph{Facility location: a survey of applications and
  methods}.\hskip 1em plus 0.5em minus 0.4em\relax Springer-Verlag New York,
  1995.

\bibitem{martella2017current}
C.~Martella, J.~Li, C.~Conrado, and A.~P. Vermeeren, ``On current crowd
  management practices and the need for increased situation awareness,
  prediction, and intervention,'' \emph{Safety Science}, vol.~91, pp. 381 --
  393, 2017.

\bibitem{handl2004evolutionary}
J.~Handl and J.~Knowles, ``Evolutionary multiobjective clustering,'' in
  \emph{International Conference on Parallel Problem Solving from
  Nature}.\hskip 1em plus 0.5em minus 0.4em\relax Springer, 2004, pp.
  1081--1091.

\bibitem{mukhopadhyay2015survey}
A.~Mukhopadhyay, U.~Maulik, and S.~Bandyopadhyay, ``A survey of multiobjective
  evolutionary clustering,'' \emph{ACM Computing Surveys}, vol.~47, no.~4, pp.
  1--46, 2015.

\bibitem{bandyopadhyay2002evolutionary}
S.~Bandyopadhyay and U.~Maulik, ``An evolutionary technique based on k-means
  algorithm for optimal clustering in rn,'' \emph{Information Sciences}, vol.
  146, no. 1-4, pp. 221--237, 2002.

\bibitem{davies1979cluster}
D.~L. Davies and D.~W. Bouldin, ``A cluster separation measure,'' \emph{IEEE
  Transactions on Pattern Analysis and Machine Intelligence}, no.~2, pp.
  224--227, 1979.

\bibitem{bezdek2013pattern}
J.~C. Bezdek, \emph{Pattern recognition with fuzzy objective function
  algorithms}.\hskip 1em plus 0.5em minus 0.4em\relax Springer Science \&
  Business Media, 2013.

\bibitem{handl2023evolutionary}
J.~Handl, M.~Garza-Fabre, and A.~Jos{\'e}-Garc{\'\i}a, ``Evolutionary
  clustering and community detection,'' in \emph{Handbook of Evolutionary
  Machine Learning}.\hskip 1em plus 0.5em minus 0.4em\relax Springer, 2023, pp.
  151--169.

\bibitem{handl2007evolutionary}
J.~Handl and J.~Knowles, ``An evolutionary approach to multiobjective
  clustering,'' \emph{IEEE Transactions on Evolutionary Computation}, vol.~11,
  no.~1, pp. 56--76, 2007.

\bibitem{garza2017improved}
M.~Garza-Fabre, J.~Handl, and J.~Knowles, ``An improved and more scalable
  evolutionary approach to multiobjective clustering,'' \emph{IEEE Transactions
  on Evolutionary Computation}, vol.~22, no.~4, pp. 515--535, 2017.

\bibitem{gama2014survey}
J.~Gama, I.~{\v{Z}}liobait{\.e}, A.~Bifet, M.~Pechenizkiy, and A.~Bouchachia,
  ``A survey on concept drift adaptation,'' \emph{ACM Computing Surveys},
  vol.~46, no.~4, pp. 1--37, 2014.

\bibitem{bifet2010moa}
A.~Bifet, G.~Holmes, B.~Pfahringer, P.~Kranen, H.~Kremer, T.~Jansen, and
  T.~Seidl, ``{MOA}: Massive online analysis, a framework for stream
  classification and clustering,'' in \emph{First Workshop on Applications of
  Pattern Analysis}.\hskip 1em plus 0.5em minus 0.4em\relax PMLR, 2010, pp.
  44--50.

\bibitem{aggarwal2003framework}
C.~C. Aggarwal, S.~Y. Philip, J.~Han, and J.~Wang, ``A framework for clustering
  evolving data streams,'' in \emph{International Conference on Very Large Data
  Bases}.\hskip 1em plus 0.5em minus 0.4em\relax Elsevier, 2003, pp. 81--92.

\bibitem{wan2008weighted}
R.~Wan, X.~Yan, and X.~Su, ``A weighted fuzzy clustering algorithm for data
  stream,'' in \emph{International Colloquium on Computing, Communication,
  Control, and Management}, vol.~1.\hskip 1em plus 0.5em minus 0.4em\relax
  IEEE, 2008, pp. 360--364.

\bibitem{aggarwal2004framework}
C.~C. Aggarwal, J.~Han, J.~Wang, and P.~S. Yu, ``A framework for projected
  clustering of high dimensional data streams,'' in \emph{International
  conference on Very large data bases}, 2004, pp. 852--863.

\bibitem{webb2016characterizing}
G.~I. Webb, R.~Hyde, H.~Cao, H.~L. Nguyen, and F.~Petitjean, ``Characterizing
  concept drift,'' \emph{Data Mining and Knowledge Discovery}, vol.~30, no.~4,
  pp. 964--994, 2016.

\bibitem{yazdani2023gnbg}
D.~Yazdani, M.~N. Omidvar, D.~Yazdani, K.~Deb, and A.~H. Gandomi, ``Gnbg: A
  generalized and configurable benchmark generator for continuous numerical
  optimization,'' \emph{arXiv preprint arXiv:2312.07083}, 2023.

\bibitem{trojanowski1999searching}
K.~Trojanowski and Z.~Michalewicz, ``Searching for optima in non-stationary
  environments,'' in \emph{Congress on Evolutionary Computation}, vol.~3, 1999,
  pp. 1843--1850.

\bibitem{peng2023evolutionary}
M.~Peng, Z.~She, D.~Yazdani, D.~Yazdani, W.~Luo, C.~Li, J.~Branke, T.~T.
  Nguyen, A.~H. Gandomi, Y.~Jin \emph{et~al.}, ``Evolutionary dynamic
  optimization laboratory: A matlab optimization platform for education and
  experimentation in dynamic environments,'' \emph{arXiv preprint
  arXiv:2308.12644}, 2023.

\bibitem{nguyen2012evolutionary}
T.~T. Nguyen, S.~Yang, and J.~Branke, ``Evolutionary dynamic optimization: A
  survey of the state of the art,'' \emph{Swarm and Evolutionary Computation},
  vol.~6, pp. 1 -- 24, 2012.

\bibitem{richter2009detecting}
H.~Richter, ``Detecting change in dynamic fitness landscapes,'' in
  \emph{Congress on Evolutionary Computation}.\hskip 1em plus 0.5em minus
  0.4em\relax IEEE, 2009, pp. 1613--1620.

\bibitem{blackwell2006multiswarm}
T.~Blackwell and J.~Branke, ``Multiswarms, exclusion, and anti-convergence in
  dynamic environments,'' \emph{IEEE Transactions on Evolutionary Computation},
  vol.~10, no.~4, pp. 459--472, 2006.

\bibitem{blackwell2008particle}
T.~Blackwell, J.~Branke, and X.~Li, ``Particle swarms for dynamic optimization
  problems,'' in \emph{Swarm Intelligence: Introduction and Applications},
  C.~Blum and D.~Merkle, Eds.\hskip 1em plus 0.5em minus 0.4em\relax Springer
  Lecture Notes in Computer Science, 2008, pp. 193--217.

\bibitem{yazdani2023species}
D.~Yazdani, D.~Yazdani, D.~Yazdani, M.~N. Omidvar, A.~H. Gandomi, and X.~Yao,
  ``A species-based particle swarm optimization with adaptive population size
  and deactivation of species for dynamic optimization problems,'' \emph{ACM
  Transactions on Evolutionary Learning and Optimization}, vol.~3, no.~4, pp.
  1--25, 2023.

\end{thebibliography}
\bibliographystyle{IEEEtran}

\newpage\appendix
\section*{Appendix}

\subsection*{Configuring Dynamic Dataset Generator}
\label{sec:parameters}

In the appendix, we explore the parameter setting of the Dynamic Dataset Generator (DDG), a tool that stands as a discrete-event-based simulation capable of replicating a diverse range of dynamic scenarios. 
The core of DDG's versatility lies in its capacity to simulate various dynamics, each governed by a distinct set of parameters. 
This multitude of parameters, while seemingly complex, is indispensable for the accurate representation and simulation of a wide range of scenarios. 
The complexity of real-world dynamics and the need for their precise emulation in a controlled environment necessitate this extensive parameterization. 
By allowing fine-grained control over each aspect of the simulation, DDG provides researchers with the flexibility to explore, test, and validate a wide range of scenarios, each with its unique characteristics and challenges.
Table~\ref{tab:DDGparameters} summarizes all the parameters used in DDG along with their suggested ranges and values.
To disable any type of dynamic in DDG, researchers can set the likelihood parameter ($\mathfrak{p}$) of that dynamic to zero. 
Additionally, to avoid changing a specific parameter, its change severity value can be set to zero.

\footnotesize
\begin{longtable}{cp{2.5cm}p{12cm}}
\caption{Summary of the parameters used in dynamic dataset generator (DDG) and their suggested values and ranges.} \label{tab:DDGparameters} \\
\midrule
Parameter & Suggested values & Description \\
\midrule
\endfirsthead

\multicolumn{3}{c}%
{{\bfseries Table \thetable\ continued from previous page}} \\
\midrule
Parameters & Suggested values & Descriptions \\
\midrule
\endhead

\midrule
\multicolumn{3}{r}{{Continued on next page}} \\ 
\endfoot

\endlastfoot
\rowcolor{gray!20}$t$ & --- & Time index representing the current state of the dynamic system.\\
$m^{(t)}$ & 3, 7, 10 or $\in\{m_\mathrm{min},\cdots,m_\mathrm{max}\}$  & Number of dynamic Gaussian components (DGCs) in DDG in the $t$th state of the dynamic system; can be constant or vary over time within $\{m_\mathrm{min},\cdots,m_\mathrm{max}\}$.\\
\rowcolor{gray!20}$[m_\mathrm{min},m_\mathrm{max}]$ & [3, 10] & Lower ($m_\mathrm{min}$) and upper ($m_\mathrm{max}$) bounds for the number of DGCs over time.\\
$d^{(t)}$ & 2, 5 or $\in\{d_\mathrm{min},\cdots,d_\mathrm{max}\}$ & Number of variables in the $t$th state of the dynamic system; scalable, but smaller values recommended due to increased complexity in higher dimensions for dynamic optimization algorithms. Constant or variable within $\{d_\mathrm{min},\cdots,d_\mathrm{max}\}$. Fixed at 2 for simulating facility location problems.\\
\rowcolor{gray!20}$[d_\mathrm{min},d_\mathrm{max}]$ & [2, 5] & Lower ($d_\mathrm{min}$) and upper ($d_\mathrm{max}$) bounds for the number of variables over time.\\
 $\vec{c}_i^{(t)}$ & $\in[Lb_j, Ub_j]$ where $j\in\{1,\cdots d^{(t)}\}$ & Center position (mean) of the $i$th DGC in the $t$th state of the dynamic system, bounded within $[Lb_j, Ub_j]$ for each dimension $j$, where $Lb_j$ and $Ub_j$ are the respective lower and upper bounds of data values.\\
\rowcolor{gray!20}$[Lb_j, Ub_j]$ & [-100, 100] & Lower ($Lb_j$) and upper ($Ub_j$) bounds in the $j$th dimension, setting limits for $\vec{c}_i^{(t)}$.
Note that while the center positions of DGCs in each dimension $j$ are bounded within $[Lb_j, Ub_j]$, it is possible for data points to be generated outside these boundaries. 
However, note that in Gaussian distributions, approximately $99.7\%$ of data are typically sampled within a range of $3\sigma_j^{(t)}$ from the mean in the $j$th dimension. 
Therefore, in extreme cases, the expected bounds for data points could extend to approximately $Lb_j - 3\sigma^{(t)}_{\mathrm{max}}$ or $Ub_j + 3\sigma^{(t)}_{\mathrm{max}}$.
Here, $\sigma_{\mathrm{max}}$ represents the largest sigma value across all dimensions and all DGCs.\\
 $\bm\sigma_i^{(t)}$ & $\in[\sigma_{\mathrm{min}}, \sigma_{\mathrm{max}}]$ & A $d^{(t)}$-dimensional vector denoting the standard deviation of the $i$th DGC in each dimension. Each $\sigma_{i,j}$ is bounded within $[\sigma_{\mathrm{min}}, \sigma_{\mathrm{max}}]$. Set all $\sigma$ values in the vector to the same value to generate spherical data.\\
\rowcolor{gray!20} $[\sigma_{\mathrm{min}}, \sigma_{\mathrm{max}}]$ & [7, 20] & Lower ($\sigma_{\mathrm{min}}=7$) and upper ($\sigma_{\mathrm{max}}=20$) bounds for $\sigma$ values.\\
$\bm\Theta_i^{(t)}$ & --- & A $d^{(t)} \times d^{(t)}$ matrix used to calculate the rotation matrix $\mathbf{R}^{(t)}_{i}$ via Algorithm~\ref{alg:RotationControlled} for the $i$th DGC. Elements on and below the principal diagonal of $\bm\Theta^{(t)}_{i}$ are zero. Set to identity matrix for no rotation.\\
\rowcolor{gray!20}$\theta^{(t)}_{i,j,k}$ &$\in [\theta_{\mathrm{min}}, \theta_{\mathrm{max}}]$& Element at the $j$th row and $k$th column of $\bm\Theta^{(t)}_{i}$, denoted $\bm\Theta^{(t)}_{i}(j,k)$, specifies the rotation angle on the $x_{j}-x_{k}$ plane, for $j < k$.\\
$[\theta_{\mathrm{min}}, \theta_{\mathrm{max}}]$ & $[-\pi, \pi]$ & Lower ($\theta_{\mathrm{min}} = -\pi$) and upper ($\theta_{\mathrm{max}} = \pi$) bounds for angle $\theta$ values.\\
\rowcolor{gray!20}$w_{i}^{(t)}$ & $\in [w_{\mathrm{min}}, w_{\mathrm{max}}]$ & Weight of the $i$th DGC in the $t$th state of the dynamic system, influencing the probability of data point generation. Weights are normalized by their sum in data generation process.\\
$[w_{\mathrm{min}}, w_{\mathrm{max}}]$ & [1, 3] & Lower ($w_{\mathrm{min}} = 1$) and upper ($w_{\mathrm{max}} = 3$) bounds for weight values. Weight values are normalized by their sum in data generation process.\\
\rowcolor{gray!20}$\tilde{s}_i$ & $\mathcal{U}(0.1,0.2)$\footnote{The notation $\mathcal{U}(l,u)$ represents a uniform distribution that generates a random number within $(l, u)$.} & Shift severity for altering $\vec{c}_i$ using Equation~\eqref{eq:CenterDyanmic1}. Used in correlation-based dynamics for frequent, minor relocations.\\
$\rho_i\in(0,1)$ & $\mathcal{U}(0.99,0.995)$ & Correlation coefficient for altering $\vec{c}_i$ using Equation~\eqref{eq:CenterDyanmic1}. A value of 0 implies fully random relocation direction, while 1 indicates completely correlated relocation along a line.\\
\rowcolor{gray!20}$\delta_{\sigma_{i,j}}$ & $\in\{-1,1\}$ & Direction factor for the $j$th dimension of $\bm\sigma^{(t)}_{i}$ in Equation~\eqref{eq:SigmaDynamic}, indicating whether $\sigma_{i,j}^{(t+1)}$ is increasing (1) or decreasing (-1).\\
$\tilde{\sigma}_i$ & $\mathcal{U}(0.05,0.1)$ & Severity of change for adjusting elements of $\sigma_i$ in Equation~\eqref{eq:SigmaDynamic}, used in correlation-based dynamics to facilitate frequent but minor changes.\\
\rowcolor{gray!20}$\delta_{w_{i}}$ & $\in\{-1,1\}$ & Direction factor for the weight of the $i$th DGC in Equation~\eqref{eq:HeightDynamic}, determining if $w_{i}^{(t+1)}$ increases (1) or decreases (-1).\\
$\tilde{w}_i$ & $\mathcal{U}(0.02,0.05)$ & Severity of change for the weight of the $i$th DGC in Equation~\eqref{eq:HeightDynamic}, employed in correlation-based dynamics for frequent, minor adjustments.\\
\rowcolor{gray!20}$\delta_{\theta_{i,j,k}}$ & $\in\{-1,1\}$ & Direction factor for the element at the $j$th row and $k$th column of $\bm{\Theta}_i^{(t)}$ in Equation~\eqref{eq:ThetaDynamic}, determining if $\theta_{i,j,k}=\bm\Theta_i^{(t+1)}(j,k)$ increases (1) or decreases (-1).\\
$\tilde{\theta}_i$ & $\mathcal{U}(\frac{\pi}{360},\frac{\pi}{180})$ & Change severity for angles in $\bm{\Theta}_i^{(t)}$ in Equation~\eqref{eq:ThetaDynamic}, used in correlation-based dynamics to facilitate frequent, minor angle adjustments.\\
\rowcolor{gray!20}$p_{l,i}$ & $\mathcal{U}(0.02,0.05)$ & Probability of changing the direction of $\delta_{w_{i}}$, $\delta_{\sigma_{i,j}}$, and $\delta_{\theta_{i,j,k}}$.\\
$\mathfrak{p}_{l,i}$ & $\mathcal{U}(0.05,0.1)$ & Probability of changing variables of the $i$th DGC at each time-tick (function evaluation in DDG) using Equations~\eqref{eq:CenterDyanmic1}, \eqref{eq:SigmaDynamic}, \eqref{eq:HeightDynamic}, and \eqref{eq:ThetaDynamic}. For deactivating gradual local changes, set $\mathfrak{p}_{l,i}=0$ for $i\in\{1,\cdots,m^{(t)}\}$.\\
\rowcolor{gray!20}$\widehat{s}$ & 2, 5 , 10 & Shift severity applied to the center (mean) position of all DGCs in Equation~\eqref{eq:Gcenter} for severe global changes.\\
$\widehat{w}$ & 0.2, 0.5 & Severity of weight change applied to all DGCs in Equation~\eqref{eq:Gheight} for severe global changes.\\
\rowcolor{gray!20}$\widehat{\sigma}$ & 2, 5 & Severity of standard deviation (width) change applied to $\bm\sigma$ of all DGCs in Equation~\eqref{eq:Gwidth} for severe global changes.\\
$\widehat{\theta}$ & $\frac{\pi}{8},\frac{\pi}{4}$ & Severity of angle change applied to $\bm\Theta$ of all DGCs in Equation~\eqref{eq:Gangle} for severe global changes.\\
\rowcolor{gray!20}$\alpha,\beta$ & 0.1 & Parameters for the Beta-distribution in Equations~\eqref{eq:Gcenter} to~\eqref{eq:Gangle}, generating heavy-tail, symmetric random values.\\
$\mathfrak{p}_g$ & 0.00005, 0.0001, 0.0002, 0.0004, 0.0008 & Probability of altering variables of all DGCs at each time-tick (function evaluation in DDG) using Equations~\eqref{eq:Gcenter} to~\eqref{eq:Gangle}. Set to 0 to deactivate severe global changes in DGC parameters.\\
\rowcolor{gray!20}$\tilde{m}$ & 1 & Severity of change for the number of DGCs, determining the quantity of DGCs to be added or removed using Equation~\eqref{eq:ChangePeakNumber}.\\
$\mathfrak{p}_m$ & 0.00005, 0.0001, 0.0002, 0.0004, 0.0008 & Probability of changing the number of DGCs at each time-tick (function evaluation in DDG) using Equation~\eqref{eq:ChangePeakNumber}. Set to 0 to deactivate changes in the number of DGCs.\\
\rowcolor{gray!20}$\tilde{d}$ & 1 & Severity of change for the number of variables in DDG, determining the quantity to be added or removed according to Equation~\eqref{eq:ChangeVariable}.\\
$\mathfrak{p}_d$ & 0.00005, 0.0001, 0.0002, 0.0004, 0.0008 & Probability of altering the number of variables at each time-tick (function evaluation in DDG) using Equation~\eqref{eq:ChangeVariable}. Set to 0 to deactivate changes in the number of variables.\\
\rowcolor{gray!20}$\kappa^{(t)}$ & 2, 5, 10 or $\in \{\kappa_\mathrm{min},\cdots,\kappa_\mathrm{max}\}$ & Number of clusters in the $t$th state. In certain scenarios, this is determined by external factors such as authority decisions or availability of facilities.\\
$[\kappa_\mathrm{min},\kappa_\mathrm{max}]$ & [2,10] & Lower ($\kappa_\mathrm{min}=2$) and upper ($\kappa_\mathrm{max}=10$) bounds for the number of clusters $\kappa$.\\
\rowcolor{gray!20}$\tilde{\kappa}$ & 1 & Severity of change for the number of clusters, determining the quantity to be added or removed as per Equation~\eqref{eq:ChangeClusterNum}.\\
$\mathfrak{p}_\kappa$ & 0.0001, 0.0002, 0.0004, 0.0008 & Probability of changing the number of clusters defined by external factors at each time-tick (function evaluation in DDG) as per Equation~\eqref{eq:ChangeClusterNum}. Set to 0 to deactivate changes in cluster numbers.\\
\rowcolor{gray!20}$|\mathbf{D}|$ & 200, 500, 1000 & Size of the dataset $\mathbf{D}$. In the initial version of DDG, the dataset size is maintained constant over time.\\
$\mathfrak{n}\%$ & 1, 2, 5 & Percentage of the dataset $\mathbf{D}$ size replaced with new data to reflect gradual changes (Section \ref{sec:LocalChanges}).\\
\rowcolor{gray!20}$\mathfrak{p}_s$ & 0.1, 0.2 & Probability of updating $\mathfrak{n}\%$ of the dataset $\mathbf{D}$ at each time-tick (function evaluation in DDG) to reflect gradual changes (Section \ref{sec:LocalChanges}). Set to 0 for scenarios in which gradual changes are disabeled.\\
$\mathrm{FE}_\mathrm{max}$ & 500,000 & Maximum number of clustering function evaluations, serving as the termination criterion.\\
   \bottomrule
\end{longtable}

\end{document}